%% file: neurips_2026.tex
\documentclass{article}

 \usepackage[preprint]{neurips_2026}

\usepackage{microtype}
\usepackage{graphicx}
\usepackage{subcaption}
\usepackage{booktabs} 
\usepackage{hyperref}

\usepackage{amsmath}
\usepackage{amssymb}
\usepackage{mathtools}
\usepackage{amsthm}
\usepackage{caption}
\usepackage{wrapfig}
\usepackage[utf8]{inputenc} 
\usepackage[T1]{fontenc}    
\usepackage{hyperref}       
\usepackage{url}            
\usepackage{multirow} 
\usepackage{booktabs}       
\usepackage{amsfonts}       
\usepackage{nicefrac}       
\usepackage{microtype}      
\usepackage{xcolor}         
\usepackage[table]{xcolor}
\usepackage{tabularx}
\usepackage{graphicx}
\usepackage{adjustbox}
\usepackage{subcaption}  
\usepackage{nicematrix}
\newcommand{\method}{Steady Thought}

\usepackage[capitalize,noabbrev]{cleveref}

\title{SteadyThought: Mitigating LLM Under-Thinking \\ via Thought-Level Preference Optimization}

\author{%
  {\bfseries
  Zhenyue Peng\textsuperscript{1,2} \quad
  Lin Yang\textsuperscript{3}\thanks{Project lead.} \quad
  Xiaoqi Ni\textsuperscript{2}} \\
  {\bfseries
  Hanzhu Chen\textsuperscript{2} \quad
  Xize Liang\textsuperscript{2} \quad
  Yanjun Wu\textsuperscript{1,2}} \\
  {\bfseries
  Rujing Wang\textsuperscript{1,2} \quad
  Kaiwen Zhou\textsuperscript{5} \quad
  Jianye Hao\textsuperscript{6} \quad
  Jie Wang\textsuperscript{4}\thanks{
Corresponding author: \texttt{jiewangx@ustc.edu.cn}.}} \\
  {\normalfont\small
  \textsuperscript{1}Hefei Institutes of Physical Science,
  Chinese Academy of Sciences, Hefei 230031, China}\\
  {\normalfont\small
  \textsuperscript{2}University of Science and Technology of China, Hefei 230026, China}\\
  {\normalfont\small
  \textsuperscript{3}Shanghai Artificial Intelligence Laboratory, Shanghai, China}\\
  {\normalfont\small
  \textsuperscript{4}MoE Key Laboratory of Brain-inspired Intelligent Perception and Cognition,}\\
  {\normalfont\small
  University of Science and Technology of China}\\
  {\normalfont\small
  \textsuperscript{5}Huawei Technologies Co., Ltd.}\\
  {\normalfont\small
  \textsuperscript{6}College of Intelligence and Computing, Tianjin University}\\
  {\normalfont\small
   \texttt{pzysa22168242@mail.ustc.edu.cn}}
}

\begin{document}

\maketitle

\begin{abstract}
 Flexible switching between reasoning trajectories (i.e., thought switching) has significantly enhanced the reasoning capabilities of Large Reasoning Models (LRMs).
However, existing models often switch excessively yet fail to sustain promising reasoning thoughts—--a phenomenon termed ``under-thinking''.
While recent efforts suppress switching to mitigate this, such over-correction may discard valuable trajectories.
To address this challenge, we propose \textbf{\underline{S}teady \underline{T}hought} (\textbf{ST}), a novel thought-level preference optimization framework.
ST formalizes under-thinking as a preference issue at switching points, forcing single-path continuations from identified points to yield high-quality reasoning. Then, ST performs thought-level preference optimization by treating the newly generated response as preferred and the original one as dis-preferred.
Experiments across multiple models and datasets show that ST effectively reduces token consumption, achieving reductions of up to 63.8\%, while maintaining strong reasoning accuracy. Further analysis suggests that ST helps models acquire a generalizable ability to reduce redundant switching across domains and languages. Code is available at
\url{https://github.com/MIRALab-USTC/SteadyThought}.
\end{abstract}

\section{Introduction}

Recent advancements in Large Reasoning Models (LRMs), such as DeepSeek-R1~\citep{deepseekai2025deepseekr1incentivizingreasoningcapability} and GPT-o1~\citep{o1}, have demonstrated remarkable capabilities in solving complex tasks. Their success is largely attributed to the emulation of human-like ``slow thinking'' processes, specifically the ability to flexibly switch between reasoning strategies—a mechanism termed  \textit{thought switching}~\citep{zeng2025simplerlzooinvestigatingtamingzero,muennighoff2025s1simpletesttimescaling,yang2025understandingahamomentsexternal,chen2025empiricalstudyelicitingimproving}. This adaptability allows models to explore diverse reasoning paths, theoretically leading to more accurate and robust inferences.

\begin{figure*}[t]
    \centering
    \begin{minipage}[b]{0.26\textwidth} 
        \begin{subfigure}[t]{\textwidth}
            \centering
            \includegraphics[width=\textwidth]{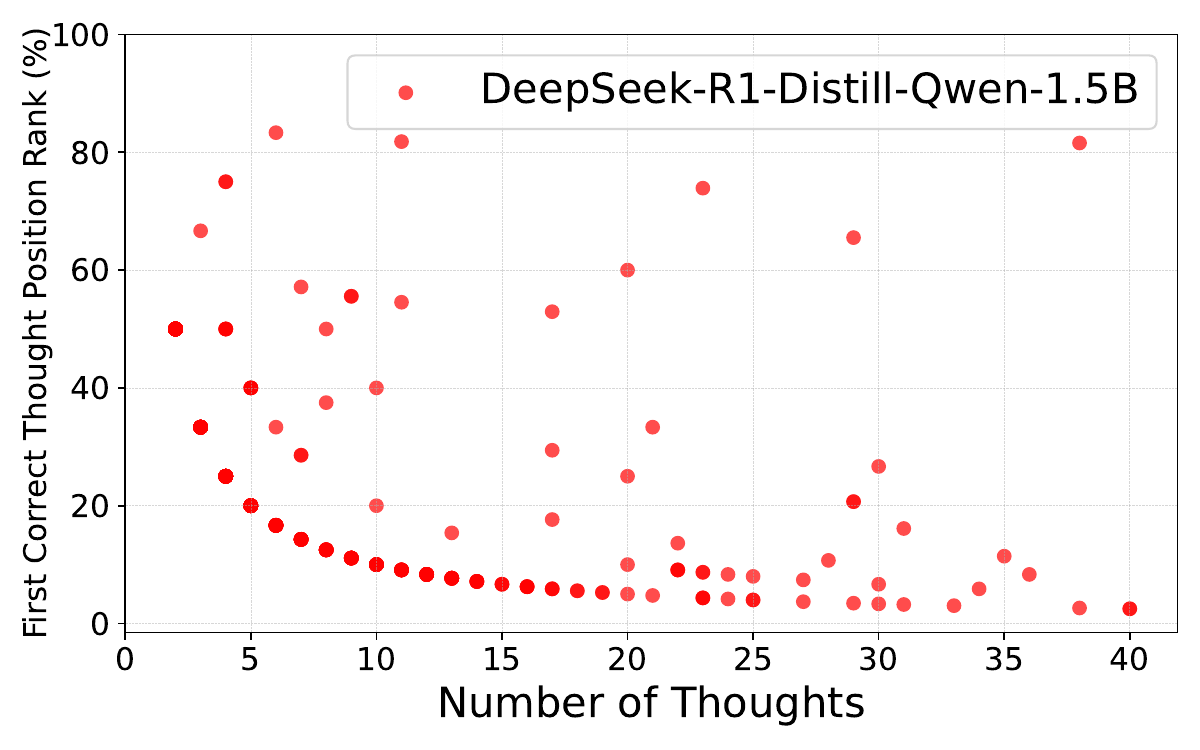}
            \caption{DeepSeek-R1-Distill-\\Qwen-1.5B on MATH-500}
            \label{r1-math-fctr}
        \end{subfigure}
        
        \vspace{0.1em} 
        
        \begin{subfigure}[t]{\textwidth}
            \centering
            \includegraphics[width=\textwidth]{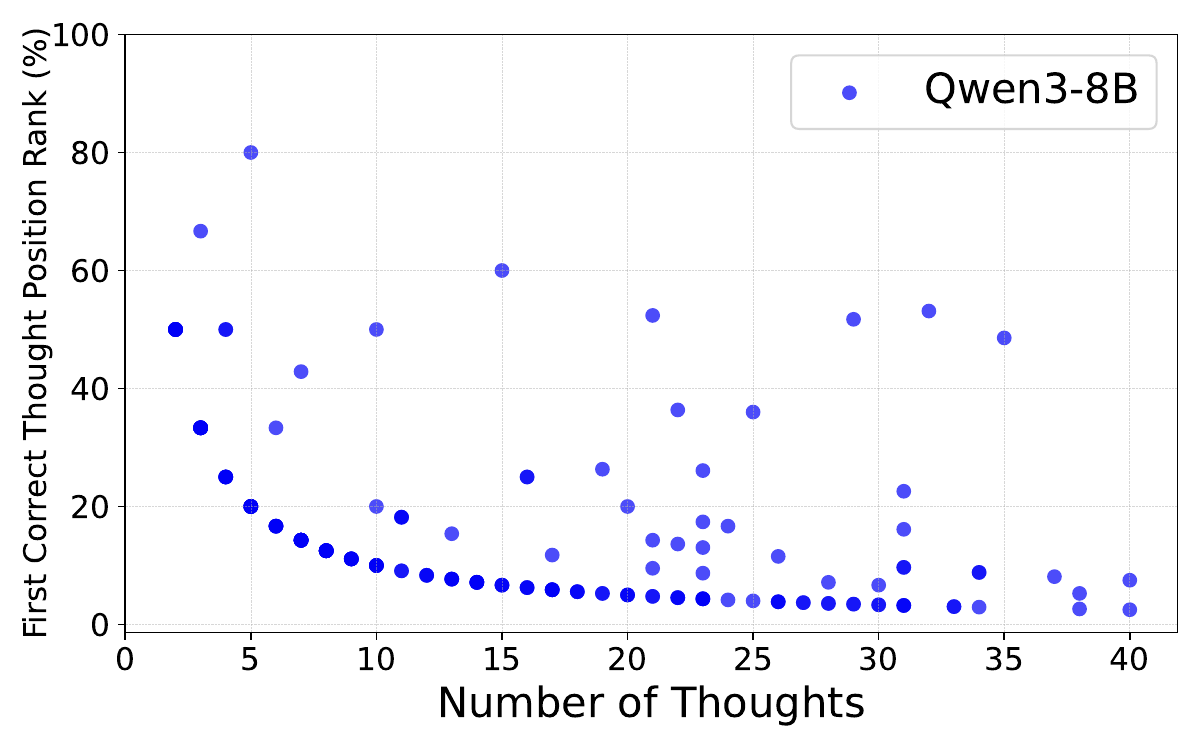}
            \caption{Qwen3-8B on AIME2024}
            \label{qwen-aime-fctr}
        \end{subfigure}
    \end{minipage}
    \hfill 
    \begin{minipage}[b]{0.69\textwidth} 
        \begin{subfigure}[t]{\textwidth}
            \centering
            \includegraphics[width=\textwidth]{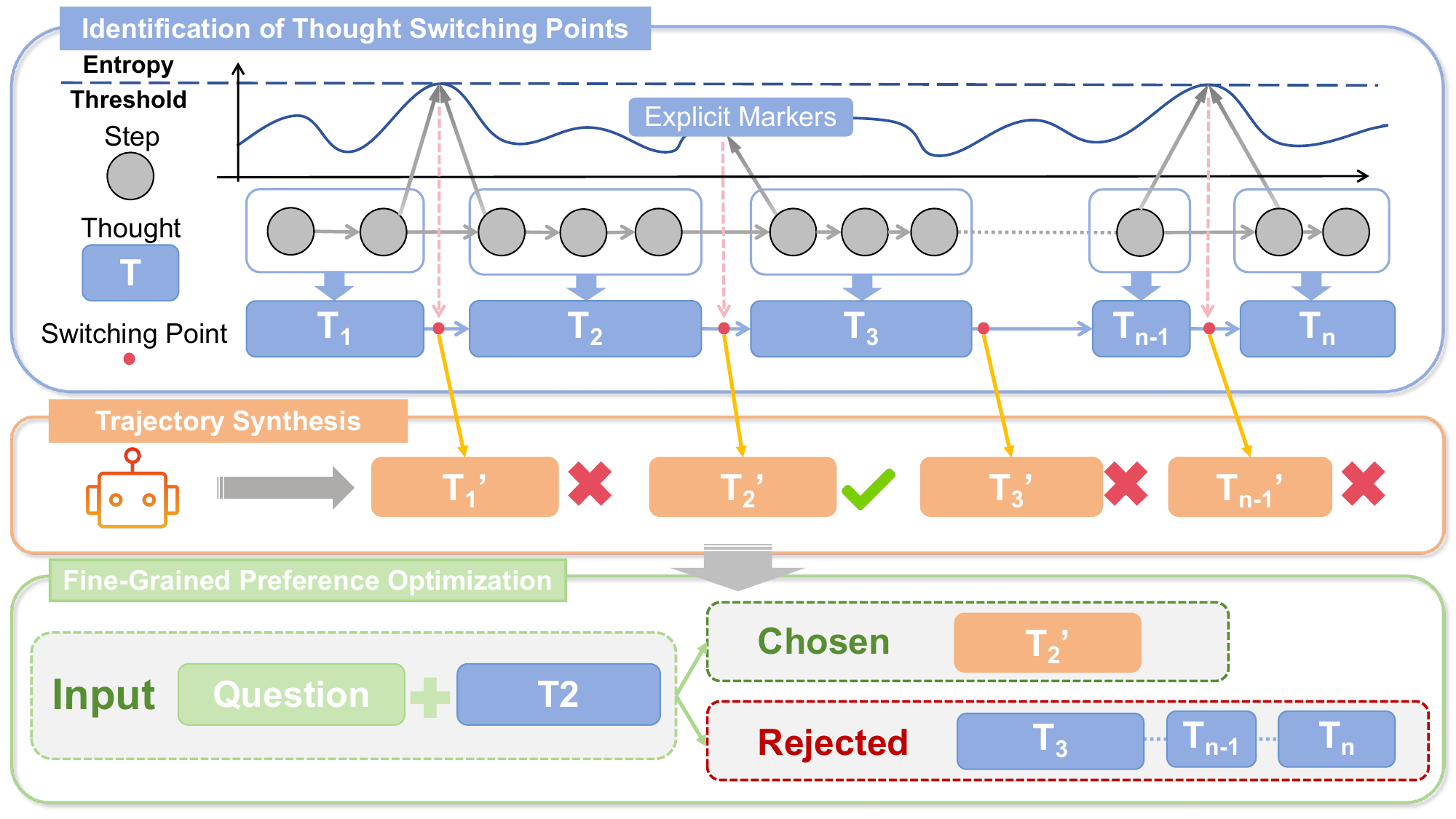} 
            \caption{Overview of Steady Thought framework}
            \label{overview}
        \end{subfigure}
    \end{minipage}

    \vspace{0.8em}
    \caption{(a) Rank of the first correct thought position for DeepSeek-R1-Distill-Qwen-1.5B on MATH-500. (b) Rank of the first correct thought position for Qwen3-8B on AIME2024. (c) Overview of the ST framework, which identifies thought switching points using two complementary signals: explicit linguistic markers and concurrent entropy spikes around step boundaries. Starting from these identified junctions, the model is guided via logit control to generate persistent reasoning continuations, effectively verifying the potential of the current thought. These synthesized paths are then used to construct fine-grained preference pairs for the STPO objective, which reinforces trajectory commitment at the exact point of divergence to effectively mitigate under-thinking and improve reasoning efficiency.}
    \vspace{-20pt}
    \label{method}
\end{figure*}

However, recent studies have shown that LRMs tend to switch thoughts too frequently~\citep{wang2025waitdontneedwait,ding2025thinkingtokenshelptrap,chen2025sealsteerablereasoningcalibration} and often fail to follow the promising thoughts---a phenomenon termed ``under-thinking''~\citep{wang2025thoughtsplaceunderthinkingo1like}.
Specifically, Figures~\ref{r1-math-fctr} and ~\ref{qwen-aime-fctr} illustrate the initial emergence of correct thoughts during the models' reasoning processes. They reveal that the models often find a reliable thought early (which can lead to the correct answer), but still proceed with numerous additional thought switches. This issue stems from the model's inability to recognize and commit to high-potential thoughts, effectively hindering the depth and coherence of reasoning. Furthermore, such erratic switching results in substantial inefficiency, wasting computational resources on unproductive generation.

To mitigate the shallow reasoning resulting from frequent thought switching, existing approaches primarily focus on indiscriminately suppressing switching behaviors.
For example, some methods~\citep{wang2025thoughtsplaceunderthinkingo1like,wang2025waitdontneedwait,ding2025thinkingtokenshelptrap} assume that the thought switching often starts with certain special tokens (e.g., ``alternatively'' or ``wait'') and discourage the model from switching by lowering the probabilities of these tokens during decoding or reducing their rewards in training. On the other hand, ~\cite{chen2025sealsteerablereasoningcalibration} take a different approach by operating in the representation space: they steer the model’s hidden states away from a learned ``switching'' vector and toward a ``reasoning'' vector, thereby suppressing switching behavior in a more structured manner.
While these methods effectively reduce the frequency of switches, they apply suppression indiscriminately, thereby restricting the model's flexibility to explore alternative paths when the current trajectory is indeed incorrect. These limitations highlight the need for a more selective mechanism---one that preserves the ability to explore new reasoning thoughts when the current trajectory is unpromising, while encouraging deeper commitment to a thought when it shows promise.

To this end, we propose \textbf{Steady Thought (ST)}, a novel thought-level preference optimization framework for mitigating under-thinking by selectively discouraging thought switching, thereby improving token efficiency without sacrificing reasoning performance.  We first formalize the under-thinking problem as a preference alignment issue at critical decision points. These points are effectively identified through a complementary approach that combines marker-based detection with the monitoring of entropy spikes at step boundaries. Subsequently, starting from the identified thought switching point, the model is guided via logit control to follow its preceding thought in a single direction without further switching until the final answer is reached. By evaluating these completions against correct answers, we obtain high-quality reasoning continuations that exhibit stable late-stage persistence, which serve as the preferred targets for optimization, and the original response---characterized by unnecessary switching—--is treated as dis-preferred. Finally, we apply a thought-level preference optimization method named \textbf{Steady Thought Preference Optimization} (STPO) to encourage the model to persist in promising reasoning directions while discouraging unnecessary switching at thought switching points.

Our main contributions are threefold:

\begin{itemize}
\setlength{\itemsep}{5pt}
\setlength{\parskip}{0pt}
\setlength{\parsep}{0pt}
\setlength{\topsep}{-10pt}
     \item We analyze the specific manifestations of the under-thinking phenomenon and formalize it as a preference optimization problem.
    \item We propose \textbf{\method} (ST), a novel thought-level preference optimization framework that encourages models to develop and stick to high-potential reasoning thoughts without compromising their flexibility to explore alternative reasoning trajectories.

    \item We validate the effectiveness of ST through extensive experiments across multiple models and reasoning benchmarks, demonstrating substantial token reductions of up to 63.8\% while maintaining strong reasoning accuracy.

\end{itemize}

\section{Background}

\subsection{Problem Formulation}
We formalize the reasoning process of a Large Reasoning Model (LRM) as a trajectory punctuated by multiple thought switching points, which naturally partitions the response $y$ to a question $x$ into a sequence of distinct thoughts:
\begin{equation}
y = (T_1, T_2, \dots, T_n),
\end{equation}
where each thought $T_i$ represents a coherent segment of reasoning. Following a sequence of reasoning explorations to reach the $i$-th thought, denoted by the prefix $\mathbf{P}_i = (\mathbf{x}, T_1, \dots, T_i)$, the policy $\pi_\theta$ defines a conditional probability distribution over all possible subsequent trajectories.

The problem of \textbf{under-thinking} arises at a critical decision point. After the model generates a \textbf{promising thought} $T_i$ (one that \textit{can} lead to a correct answer), it faces a choice. Within the set of possible next trajectories, we are interested in two specific types:
\begin{itemize}
\setlength{\itemsep}{3pt}
\setlength{\parskip}{0pt}
\setlength{\parsep}{0pt}
\setlength{\topsep}{0pt}
    \item The \textbf{Commit Trajectory ($\mathbf{\tau_c}$)}: The ideal trajectory where the model commits to and correctly completes the promising thought $T_i$. We denote this completed sequence as $T_i'$.
    \item The \textbf{Switch Trajectory ($\mathbf{\tau_s}$)}: The suboptimal trajectory observed in the data, where the model abandons $T_i$ and switches to a new, often wasteful line of reasoning $(T_{i+1}, \dots, T_n)$.
\end{itemize}

Our objective is to align the model's policy $\pi_\theta$ with the preference for the commit trajectory over the switch trajectory, i.e., $\mathbf{\tau_c} \succ \mathbf{\tau_s}$. To formalize this, we define a latent \textbf{Steadiness Score} $S_\pi(\mathbf{\tau} | \mathbf{P}_i)$, which quantifies the policy's inclination towards a specific trajectory $\mathbf{\tau}$ given the prefix $\mathbf{P}_i$. The preference $\mathbf{\tau_c} \succ \mathbf{\tau_s}$ implies that the score of the commit trajectory should be higher.

Following the Bradley-Terry model, the probability of this preference can be expressed through the difference in scores:
\begin{equation}
    P(\mathbf{\tau_c} \succ \mathbf{\tau_s} | \mathbf{P}_i) = \sigma(S_\pi(\mathbf{\tau_c}|\mathbf{P}_i) - S_\pi(\mathbf{\tau_s}|\mathbf{P}_i)),
\end{equation}
where $\sigma$ is the sigmoid function. This abstract scoring function provides a powerful new lens through which to view the problem. Crucially, it can be directly instantiated by the LRM's own log-probabilities. Preference optimization methods like Direct Preference Optimization (DPO)~\citep{rafailov2024direct}  and Simple Preference Optimization (SimPO)~\citep{meng2024simposimplepreferenceoptimization} allow us to directly optimize the policy $\pi_\theta$ such that its implicit scoring function, where $S_\pi(\mathbf{\tau}|\mathbf{P}_i) := \log \pi_\theta(\mathbf{\tau}|\mathbf{P}_i)$, satisfies the desired preference relationship. This provides a principled way to train the model to be a more steadfast and effective reasoner.

\subsection{Preference Optimization Methods}
DPO significantly advanced the alignment of language models with human preferences by leveraging the Bradley-Terry preference model to bypass the complex reward modeling and policy optimization of traditional reinforcement learning. However, DPO is sensitive to response-length differences because its sequence-level log-probability ratios accumulate token-level contributions, making the preference signal dependent on response length. SimPO addresses this limitation by introducing a length-normalized reward with a target margin, effectively mitigating length bias without requiring a reference model. Given the preference data $\mathcal{D} = \{(\mathbf{x}, \mathbf{y}_w, \mathbf{y}_l)\}$, where $\mathbf{x}$ is the input prompt, $\mathbf{y}_w$ is the preferred response, and $\mathbf{y}_l$ is the dispreferred response. SimPO optimizes the following objective:
\begin{equation}
\begin{aligned}
\mathcal{L}_{\mathrm{SimPO}}(\pi_\theta)
= - \mathbb{E}_{(x,y_w,y_l)\sim\mathcal{D}}
\left[
\log \sigma
\left(
\frac{\beta}{|y_w|}\log \pi_\theta(y_w \mid x)
-
\frac{\beta}{|y_l|}\log \pi_\theta(y_l \mid x)
-
\gamma
\right)
\right].
\end{aligned}
\end{equation}
where $\sigma$ is the sigmoid function, with $\beta$ as a temperature parameter controlling sensitivity to preference differences and $\gamma$ as the target reward margin for the model to achieve.

\section{Method}

We propose Steady Thought (ST), a preference optimization framework designed to encourage reasoning persistence in promising reasoning thoughts. As illustrated in Figure~\ref{overview}, ST first identifies thought switching points and then synthesizes persistent trajectories from these points. Finally, it applies fine-grained preference optimization to reinforce commitment to promising thoughts and reduce unnecessary switching.

\subsection{Identification of Thought Switching Points}
\label{sec:3.1}

To identify points at which a reasoning trajectory changes direction, we propose a complementary dual-channel detection framework. We first partition each model response into a sequence of candidate steps using the structural delimiter ``\texttt{\textbackslash n\textbackslash n}''. Each step boundary is then examined using two complementary signals:

\begin{itemize}
\setlength{\itemsep}{2pt}
\setlength{\parskip}{0pt}
\setlength{\parsep}{0pt}
\setlength{\topsep}{0pt}
\item \textbf{Explicit Linguistic Markers:} The subsequent step begins with a predefined phrase that explicitly signals a change in reasoning direction. Representative markers include ``alternatively'' and ``another way,'' with the full list provided in Appendix~\ref{Markers}.

\item \textbf{Boundary Entropy Spikes:} High entropy values (exceeding a threshold $\tau$) are observed concurrently at the terminal tokens of the preceding step and the initial tokens of the next step. At each token position, the entropy of the model's predictive distribution is computed as
\begin{equation}
H(P) = -\sum_{v \in \mathcal{V}} P(v) \log P(v),
\end{equation}
where $\mathcal{V}$ denotes the vocabulary and $P(v)$ is the predicted probability of token $v$ at that position.

\end{itemize}

To operationalize entropy-based switching detection, we calibrate an entropy threshold $\tau$ using responses sampled from the training pool. For each model, we select the lowest candidate threshold at which at least 80\% of the detected boundaries correspond to genuine switches in reasoning direction, thereby balancing detection reliability and boundary coverage. The calibrated threshold is then fixed and applied to the full training set for thought switching point detection. Given this threshold, a candidate step boundary satisfies the entropy criterion when predictive entropy simultaneously exceeds $\tau$ on both sides of the boundary, at the end of the preceding step and the beginning of the subsequent step. Further details of the calibration procedure and threshold sensitivity analysis are provided in Appendix~\ref{thresholds}.

A step boundary is identified as a \textbf{thought switching point} if either the linguistic-marker condition or the entropy criterion is satisfied. Each switching point marks the beginning of a new thought, and subsequent steps are grouped into the same thought until the next switching point is encountered. This process partitions the original response $\mathbf{y}$ into a structured sequence of thoughts:
\begin{equation}
\mathbf{y} = (T_1, T_2, \dots, T_n).
\end{equation}
The rationale for this dual-channel approach lies in the complementary nature of linguistic and stochastic signals. Explicit keywords are highly effective at capturing universal and structured switching points across different models. In contrast, some studies have indicated that high entropy of tokens serves as a signal for latent uncertainty in the model's predictive distribution, which usually occurs when the model is re-planning or exploring a new reasoning trajectory~\citep{wang20258020rulehighentropyminority,cui2025entropymechanismreinforcementlearning}. This allows for the detection of model-specific or problem-specific implicit switches that lack formal transition words. For example, when solving a mathematical problem concerning properties of regular $n$-gons under permutations, the model reaches an impasse after an extended but inconclusive qualitative analysis of inscribed angles and arcs. It then pivots to a coordinate-based analytical approach, stating, ``Let me think about the coordinates...''. Although this transition does not contain any of our predefined linguistic markers, it is accompanied by a pronounced entropy spike around the step boundary, reflecting the shift from qualitative geometric reasoning to a different analytical strategy. By combining the two channels, our method captures both explicit linguistic shifts and implicit uncertainty-associated transitions. We provide an ablation study of the two detection channels in Appendix~\ref{Switching-Point}.

\subsection{Trajectory Synthesis at Switching Points}
\label{sec:3.2}

Once a thought switching point is identified, we must determine whether the model should have persisted with the current reasoning thought or transitioned to an alternative one, as exhibited in the original response. We evaluate the potential of each segmented thought $T_i$ by generating a continuation from the question $Q$ and the current thought itself. Since switching points are used to partition the trajectory into localized reasoning units, this setup allows us to directly assess whether $T_i$ can support a correct continuation. It also reduces the context length during trajectory synthesis, allowing more thought-level instances to be retained under limited memory and context budgets.

To ensure the model remains committed to the current trajectory, we apply logit control during the decoding phase. Specifically, we sharply decrease the logit values of transition tokens that typically trigger a switch, such as ``wait'', ``but'' or ``however''. By driving the selection probability of these tokens toward zero, we force the model to pursue the current line of reasoning without further deviation until a final answer is generated. This synthesized completion is denoted as:
\begin{equation}
T_i' \sim \pi_\theta(\cdot \mid Q, T_i),
\end{equation}
where $T_i'$ represents the model's self-generated reasoning path under the constraint of reasoning persistence. We retain $T_i$ as a promising thought only if the corresponding $T_i'$ reaches the correct final answer; otherwise, it is discarded. Further analysis of the synthesis cost and data utilization is provided in Appendix~\ref{Construction}.

\subsection{Fine-Grained Preference Optimization}

A significant limitation of holistic preference optimization is that it treats entire reasoning chains as monolithic blocks. For complex multi-step problems, an incorrect response often contains a substantial sequence of valid initial reasoning. Rejecting such a response in its entirety discards these valuable correct segments and provides a noisy, unfocused learning signal.

To address this, our approach provides more granular supervision by focusing on the critical switching points where the reasoning diverges. We aim to teach the model to commit to a promising thought once it has been identified, rather than abandoning it. Specifically, an answer $\mathbf{y}$ can be decomposed into a sequence of thoughts $\mathbf{y} = (T_1, \dots, T_n)$. When we identify a promising thought $T_i$, the optimization then hinges on the trajectory the model takes from this point forward, conditioned on both the original question $Q$ and the thought $T_i$ itself.

The preference pair is constructed based on the continuations from this shared context:
\begin{align*}
    \text{Chosen:} \quad & \mathbf{y}_{w} = T_i' 
    & \text{\small (The full, correct completion)} \\[0.5ex]
    \text{Rejected:} \quad & \mathbf{y}_{l} = ( T_{i+1},T_{i+2}, \dots, T_n ) 
    & \text{\small (The subsequent wasteful thoughts)}
\end{align*}
Inspired by the reference-free and length-normalized objective of SimPO, the STPO loss is formulated as:
\begin{equation}
\small 
\begin{aligned}
\mathcal{L}_{\text{STPO}}(\pi_\theta) = - \mathbb{E}_{(Q, T_i, \mathbf{y}_w, \mathbf{y}_l) \sim \mathcal{D}} \Bigl[ \log \sigma \Bigl( & \frac{\beta}{|\mathbf{y}_w|} \log \pi_\theta(\mathbf{y}_w|Q, T_i) 
- \frac{\beta}{|\mathbf{y}_l|} \log \pi_\theta(\mathbf{y}_l|Q, T_i) - \gamma \Bigr) \Bigr].
\end{aligned}
\end{equation}
This formulation provides a crucial distinction from conventional preference tuning. The learning signal, embodied by the conditional log-probabilities $\log \pi_\theta(\cdot|Q, T_i)$, is applied directly at the point of divergence. It forces the model to learn not just what a good final answer looks like, but to recognize and commit to a promising intermediate thought. This targeted, thought-level supervision is key to effectively mitigating the model's tendency to ``under-think'' and abandon viable reasoning paths.

\section{Experiments}

\subsection{Datasets}
We selected the omni-math~\citep{gao2024omnimathuniversalolympiadlevel} dataset as the source of our training data. This dataset contains thousands of problems at the level of the International Mathematical Olympiad, which are categorized by difficulty. We sampled problems from various difficulty levels and used a target model (e.g., Qwen3-8B) to perform inference and generate responses.

We evaluate our method on four datasets: 
(1) MATH-500~\citep{hendrycks2021measuringmathematicalproblemsolving}, which contains 500 math problems across diverse domains such as algebra, geometry, and number theory; 
(2) AIME 2024~\citep{maa-aime-2024}, a challenging math competition benchmark consisting of 30 problems from the American Invitational Mathematics Examination; 
(3) GSM8K~\citep{cobbe2021trainingverifierssolvemath}, which contains 1,319 grade-school math word problems requiring multi-step reasoning; and 
(4) LiveCode~\citep{jain2024livecodebenchholisticcontaminationfree}, which evaluates coding ability with 400 programming problems collected from competitive programming platforms. Since our training data is drawn only from mathematical datasets, we use LiveCode as an out-of-distribution benchmark to evaluate generalization beyond math.

\subsection{Baseline and Metrics}
We compare Steady Thought against four base models (DeepSeek-R1-Distill-Qwen-1.5B~\citep{deepseekai2025deepseekr1incentivizingreasoningcapability}, Qwen3-8B~\citep{qwen3-2024}, DeepSeek-R1-Distill-Qwen-14B~\citep{deepseekai2025deepseekr1incentivizingreasoningcapability} and Qwen3-32B~\citep{qwen3-2024}), as well as four test-time efficiency methods, namely NoThink, NOWAIT, SEAL and TIP.
\begin{itemize}
\setlength{\itemsep}{3pt}
\setlength{\parskip}{0pt}
\setlength{\parsep}{0pt}
\setlength{\topsep}{-2pt}
\item NoThink: ~\cite{ma2025reasoningmodelseffectivethinking} make the model skip the thinking process and output the response directly by adding the \verb|</think>| tag after the \verb|<think>| tag in the prompt.
\item NOWAIT: ~\cite{wang2025waitdontneedwait} reduce the logits values of certain keywords that represent reflection (e.g., ``wait'') during decoding, making them almost impossible to output.
\item SEAL: ~\cite{chen2025sealsteerablereasoningcalibration} first collect multiple responses and categorize them into three types of thoughts: execution, reflection, and transition. A differential vector S is derived in the latent space between execution vectors and the combination of reflection/transition vectors. During decoding, the hidden states are modified as \(\tilde{H} = H + \alpha \cdot S\), biasing the hidden layer toward execution-oriented thoughts.
\item TIP: ~\cite{wang2025thoughtsplaceunderthinkingo1like} reduce the logits of switching tokens within a local decoding window and restores their original probabilities outside the window until the next thought is reached, which encourages longer intervals between switching tokens.
\end{itemize}
We report the accuracy and average token count for each task. We average over eight test runs for AIME 2024 and two runs for LiveCode. Since MATH-500 and GSM8K contain larger test sets and exhibit relatively stable model outputs, we use a single test run for these two datasets.

\subsection{Main Results}

\begin{table*}[h]
\centering
\small 
\renewcommand{\arraystretch}{0.90} 
\caption{Experimental results on four large reasoning models.}
\label{main}
\begin{adjustbox}{max width=\textwidth}
\begin{tabular}{@{}lcccccccccc@{}}
\toprule
\multirow{2}{*}{\textbf{Method}}
& \multicolumn{2}{c}{\textbf{MATH-500}} 
& \multicolumn{2}{c}{\textbf{AIME 2024}} 
& \multicolumn{2}{c}{\textbf{GSM8K}}
& \multicolumn{2}{c}{\textbf{LiveCode}}
& \multicolumn{2}{|c}{\textbf{Overall}} \\
& Acc(\%)$\uparrow$ & Tokens$\downarrow$  
& Acc(\%)$\uparrow$ & Tokens$\downarrow$  
& Acc(\%)$\uparrow$ & Tokens$\downarrow$  
& Acc(\%)$\uparrow$ & Tokens$\downarrow$
& Acc(\%)$\uparrow$ & Tokens$\downarrow$\\
\midrule
\multicolumn{11}{l}{{\cellcolor[rgb]{0.957,0.957,0.957}}\textit{\textbf{DeepSeek-R1-Distill-Qwen-1.5B}}} \\
\textit{Vanilla} & 82.0 & 4385 & 27.5 & 11273 & 81.9 & 1448 & 30.3 & 9623 & \multicolumn{1}{|c}{55.43} & 6682 \\
\textit{NoThink} & 65.8 & 749 & 8.7 & 3185 & 53.6 & 263 & 20.7 & 813 & \multicolumn{1}{|c}{37.20 (-18.23)} & 1252 (-81.3\%) \\
\textit{NOWAIT} & 81.0 & 2590 & 22.1 & 6941 & 79.9 & 727 & 26.8 & 5122 & \multicolumn{1}{|l}{52.45 (-2.98)} & 3845 (-42.4\%) \\
\textit{SEAL} & 82.6 & 3252 & 25.4 & 9120 & 79.7 & 860 & 29.5 & 7948 & \multicolumn{1}{|l}{54.30 (-1.13)} & 5295 (-20.8\%) \\
\textit{TIP} & 82.2 & 4462 & 30.4 & 11037 & 81.6 & 1292 & 29.5 & 9574 & \multicolumn{1}{|l}{55.93 (+0.50)} & 6591 (-1.3\%) \\
\textit{\textbf{Steady Thought}} & 83.6 & 2691 & 30.8 & 7577 & 81.5 & 1118 & 33.2 & 7335 & \multicolumn{1}{|l}{57.28 (+1.85)} & 4680 (-30.0 \%) \\
\midrule
\multicolumn{11}{l}{{\cellcolor[rgb]{0.957,0.957,0.957}}\textit{\textbf{Qwen3-8B}}} \\
\textit{Vanilla} & 91.4 & 4724 & 62.1 & 10895 & 95.6 & 1759 & 71.8 & 7112 & \multicolumn{1}{|c}{80.23} & 6122 \\
\textit{NoThink} & 85.2 & 933 & 25.8 & 3504 & 93.6 & 289 & 45.6 & 584 & \multicolumn{1}{|l}{62.55 (-17.68)} & 1327 (-78.3\%) \\
\textit{NOWAIT} & 92.8 & 3327 & 55.8 & 9371 & 95.5 & 1067 & 80.3 & 5285 & \multicolumn{1}{|l}{81.10 (+0.87)} & 4762 (-22.2\%) \\
\textit{SEAL} & 92.2 & 4034 & 58.8 & 10372 & 95.9 & 1421 & 83.4 & 6414 & \multicolumn{1}{|l}{82.58 (+2.35)} & 5560 (-9.2\%) \\
\textit{TIP} & 92.6 & 4516 & 58.3 & 10892 & 95.9 & 1705 & 81.1 & 7072 & \multicolumn{1}{|l}{81.98 (+1.75)} & 6046 (-1.2\%) \\
\textit{\textbf{Steady Thought}} & 93.4 & 2139 & 64.6 & 7640 & 95.7 & 637 & 83.2 & 5631 & \multicolumn{1}{|l}{84.23 (+4.00)} & 4011 (-34.5 \%) \\
\midrule
\multicolumn{11}{l}{{\cellcolor[rgb]{0.957,0.957,0.957}}\textit{\textbf{DeepSeek-R1-Distill-Qwen-14B}}} \\
\textit{Vanilla} & 93.6 & 3349 & 60.4 & 8974 & 94.8 & 894 & 70.1 & 6789 & \multicolumn{1}{|c}{79.73} & 5001 \\
\textit{NoThink} & 41.8 & 824 & 27.1 & 3279 & 90.1 & 256 & 44.0 & 708 & \multicolumn{1}{|l}{50.75 (-28.98)} & 1266 (-74.7\%) \\
\textit{NOWAIT} & 89.8 & 2255 & 49.2 & 7030 & 93.7 & 692 & 70.8 & 4921 & \multicolumn{1}{|l}{75.88 (-3.85)} & 3724 (-25.5\%) \\
\textit{SEAL} & 92.6 & 3253 & 60.8  & 8831 & 94.7 & 880 & 75.1 & 6706 & \multicolumn{1}{|l}{80.80 (+1.07)} & 4917 (-1.7\%) \\
\textit{TIP} & 93.0 & 3098 & 62.5 & 9060 & 94.7 & 875 & 73.5 & 6958 & \multicolumn{1}{|l}{80.93 (+1.20)} & 4998 (-0.1\%) \\
\textit{\textbf{Steady Thought}} & 93.4 & 2457 & 64.6 & 7535 & 95.3 & 791 & 74.8 & 5361 & \multicolumn{1}{|l}{82.03 (+2.30)} & 4036 (-19.3\%) \\
\midrule
\multicolumn{11}{l}{{\cellcolor[rgb]{0.957,0.957,0.957}}\textit{\textbf{Qwen3-32B}}} \\
\textit{Vanilla} & 94.0 & 4000 & 67.1 & 10087 & 95.7 & 1276 & 82.6 & 7361 & \multicolumn{1}{|c}{84.85} & 5681 \\
\textit{NoThink} & 93.2 & 2913 & 64.6 & 9772 & 93.6 & 479 & 58.4 & 272 & \multicolumn{1}{|l}{77.45 (-7.40)} & 3359 (-40.9\%) \\
\textit{NOWAIT} & 93.8 & 2982 & 67.5 & 8772 & 95.9 & 934 & 85.4 & 5990 & \multicolumn{1}{|l}{85.65 (+0.80)} & 4670 (-17.7\%) \\
\textit{SEAL} & 94.8 & 3470 & 73.3  & 9461 & 96.4 & 1078 & 85.9 & 6442 & \multicolumn{1}{|l}{87.60 (+2.75)} & 5112 (-10.0\%) \\
\textit{TIP} & 94.2 & 3951 & 67.5 & 10195 & 95.9 & 1271 & 82.0 & 7478 & \multicolumn{1}{|l}{84.90 (+0.05)} & 5723 (+0.8\%) \\
\textit{\textbf{Steady Thought}} & 94.8 & 2819 & 73.8 & 8225 & 96.6 & 844 & 85.9 & 6193 & \multicolumn{1}{|l}{87.78 (+2.93)} & 4520 (-20.4\%) \\
\bottomrule
\end{tabular}
\end{adjustbox}
\vspace{-3pt}
\end{table*}

As shown in Table~\ref{main}, Steady Thought consistently reduces reasoning token consumption across four models with different scales and architectures. Compared with the vanilla models, ST reduces the average token consumption by 30.0\%, 34.5\%, 19.3\%, and 20.4\% on the 1.5B, 8B, 14B, and 32B models, respectively. Importantly, these efficiency gains do not come at the expense of reasoning performance: ST achieves average accuracy improvements of 1.85, 4.00, 2.30, and 2.93 percentage points across the four models. Unlike methods that indiscriminately suppress thinking or switching behaviors, ST selectively discourages unnecessary switching, leading to more efficient reasoning without sacrificing overall performance.

Notably, LiveCode serves as an OOD benchmark because all models are trained only on mathematical data. On this coding benchmark, ST reduces token usage by 23.8\%, 20.8\%, 21.0\%, and 15.9\% across the four models, respectively. These efficiency gains are accompanied by accuracy improvements of 2.9, 11.4, 4.7, and 3.3 percentage points. The consistent improvements on an unseen domain indicate that the benefits of ST extend beyond the mathematical training distribution, suggesting that the learned behavior can generalize across reasoning domains. A trajectory comparison for the same task before and after ST training is provided in Appendix~\ref{example}.

\subsection{Broader Generality Across Diverse Reasoning Domains}

To further evaluate the generalization ability of ST beyond mathematical and coding domains, we still use only mathematical data for training and introduce three out-of-distribution datasets  from broader, less logic-intensive domains for evaluation: MuSR~\citep{sprague2023musr} for multi-step soft reasoning, StrategyQA~\citep{geva2021did} for commonsense reasoning, and MedQA-USMLE~\citep{jin2021disease} for medical question answering. We report the average over four evaluation runs for each dataset.\\
As shown in Table~\ref{broader}, ST substantially reduces token consumption across the three OOD datasets. For DeepSeek-R1-Distill-Qwen-1.5B, the average token consumption decreases by 60.2\%, accompanied by an average accuracy improvement of 3.07 percentage points. For Qwen3-8B, ST reduces the average output length by 34.3\% while improving the average accuracy by 1.34 percentage points. These results indicate that ST can still transfer to broader reasoning scenarios such as commonsense understanding, multi-hop reasoning, and domain-specific question answering. This further suggests that ST enables the model to acquire a transferable capability, rather than merely memorizing superficial patterns from the mathematical training data.
\begin{table*}[t]
\centering
\small 
\renewcommand{\arraystretch}{0.90} 
\caption{Experimental results on diverse OOD domains}
\label{broader}
\begin{adjustbox}{max width=\textwidth}
\begin{tabular}{@{}lcccccc|cc@{}}
\toprule
\multirow{2}{*}{\textbf{Method}}
& \multicolumn{2}{c}{\textbf{MuSR}} 
& \multicolumn{2}{c}{\textbf{StrategyQA}} 
& \multicolumn{2}{c}{\textbf{MedQA-USMLE}}
& \multicolumn{2}{|c}{\textbf{Overall}} \\
& Acc(\%)$\uparrow$ & Tokens$\downarrow$  
& Acc(\%)$\uparrow$ & Tokens$\downarrow$  
& Acc(\%)$\uparrow$ & Tokens$\downarrow$  
& Acc(\%)$\uparrow$ & Tokens$\downarrow$\\
\midrule
\multicolumn{9}{l}{{\cellcolor[rgb]{0.957,0.957,0.957}}\textit{\textbf{DeepSeek-R1-Distill-Qwen-1.5B}}} \\
\textit{Vanilla} & 40.7 & 2590 & 58.7 & 538 & 29.3 & 2077  & \multicolumn{1}{|c}{42.90} & 1735 \\
\textit{NoThink} & 31.2 & 248 & 56.0 & 83 & 27.2 & 290 & \multicolumn{1}{|l}{38.13 (-4.77)} & 207 (-88.1\%) \\
\textit{NOWAIT} & 41.5 & 578 & 59.3 & 433 & 28.6 & 955  & \multicolumn{1}{|l}{43.13 (+0.23)} & 655 (-62.2\%) \\
\textit{SEAL} & 21.8 & 1424 & 57.7 & 469 & 24.4 & 1302  & \multicolumn{1}{|l}{34.63 (-8.27)} & 1065 (-38.6\%) \\
\textit{TIP} & 39.8 & 2518 & 58.4 & 526 & 28.8 & 1901  & \multicolumn{1}{|l}{42.33 (-0.57)} & 1648 (-5.0\%) \\
\textit{\textbf{Steady Thought}} & 46.3 & 592 & 59.7 & 414 & 31.9 & 1068  & \multicolumn{1}{|l}{45.97 (+3.07)} & 691 (-60.2\%) \\
\midrule
\multicolumn{9}{l}{{\cellcolor[rgb]{0.957,0.957,0.957}}\textit{\textbf{Qwen3-8B}}} \\
\textit{Vanilla} & 58.9 & 2176 & 79.7 & 622 & 79.9 & 1936  & \multicolumn{1}{|c}{72.83} & 1578 \\
\textit{NoThink} & 55.7 & 818 & 78.1 & 313 & 61.4 & 741  & \multicolumn{1}{|l}{65.07 (-7.76)} & 624 (-60.5\%) \\
\textit{NOWAIT} & 60.7 & 1243 & 82.6 & 438 & 78.0 & 1341  & \multicolumn{1}{|l}{73.77 (+0.94)} & 1007 (-36.2\%) \\
\textit{SEAL} & 24.3 & 1906 & 57.4 & 545 & 24.8 & 1761  & \multicolumn{1}{|l}{35.50 (-37.33)} & 1404 (-11.0\%) \\
\textit{TIP} & 59.9 & 2147 & 79.8 & 612 & 80.2 & 1953  & \multicolumn{1}{|l}{73.30 (+0.47)} & 1570 (-0.5\%) \\
\textit{\textbf{Steady Thought}} & 59.3 & 1433 & 83.3 & 382 & 79.9 & 1295  & \multicolumn{1}{|l}{74.17 (+1.34)} & 1036 (-34.3\%) \\
\bottomrule
\end{tabular}
\end{adjustbox}
\vspace{-10pt}
\end{table*}

\subsection{In-depth Analysis and Ablation Studies}

\subsubsection{Analysis of Thought Switching}

\begin{figure*}[t!]
    \centering
    
    \begin{subfigure}[b]{0.24\textwidth}
        \centering
        \includegraphics[width=\linewidth]{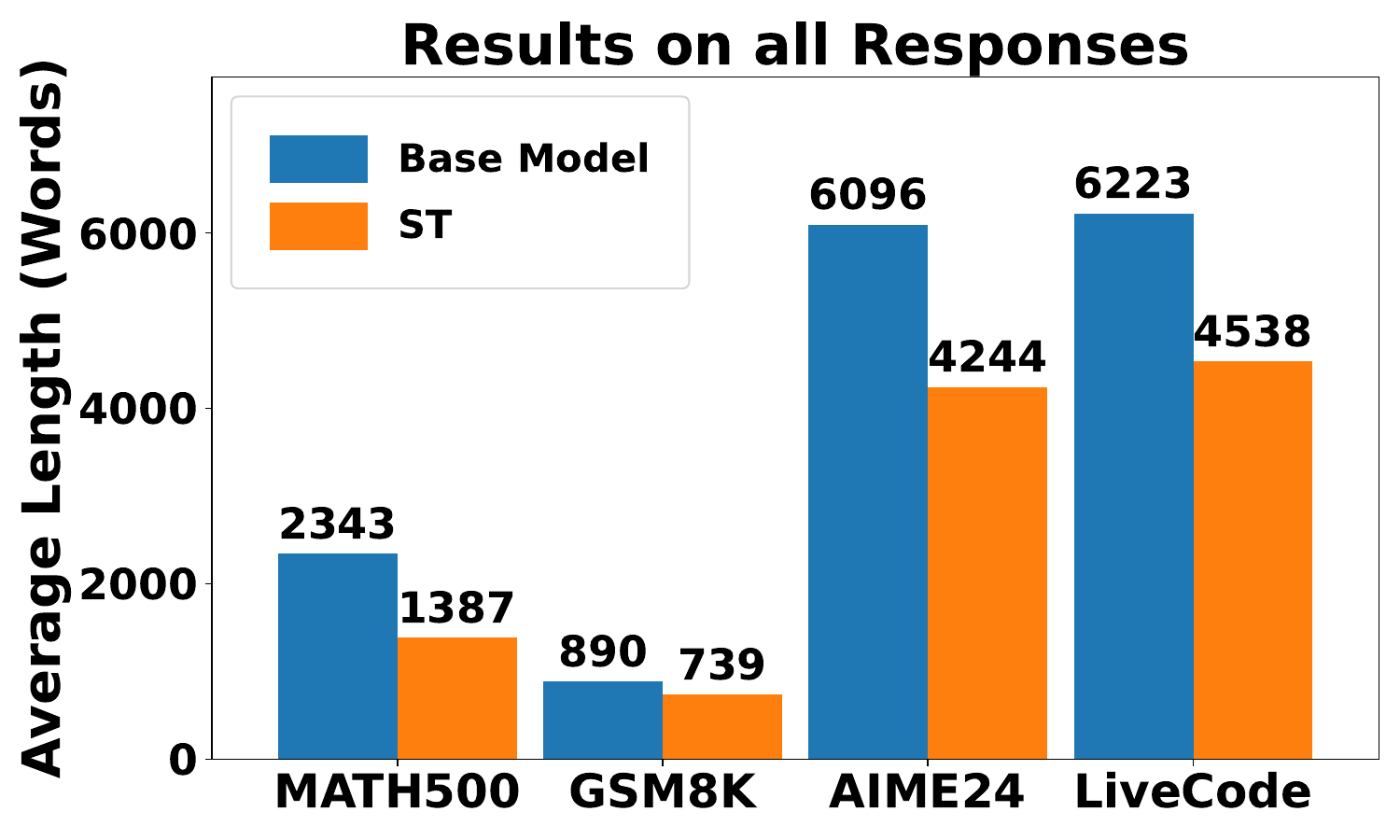}
    \end{subfigure}\hfill
    \begin{subfigure}[b]{0.24\textwidth}
        \centering
        \includegraphics[width=\linewidth]{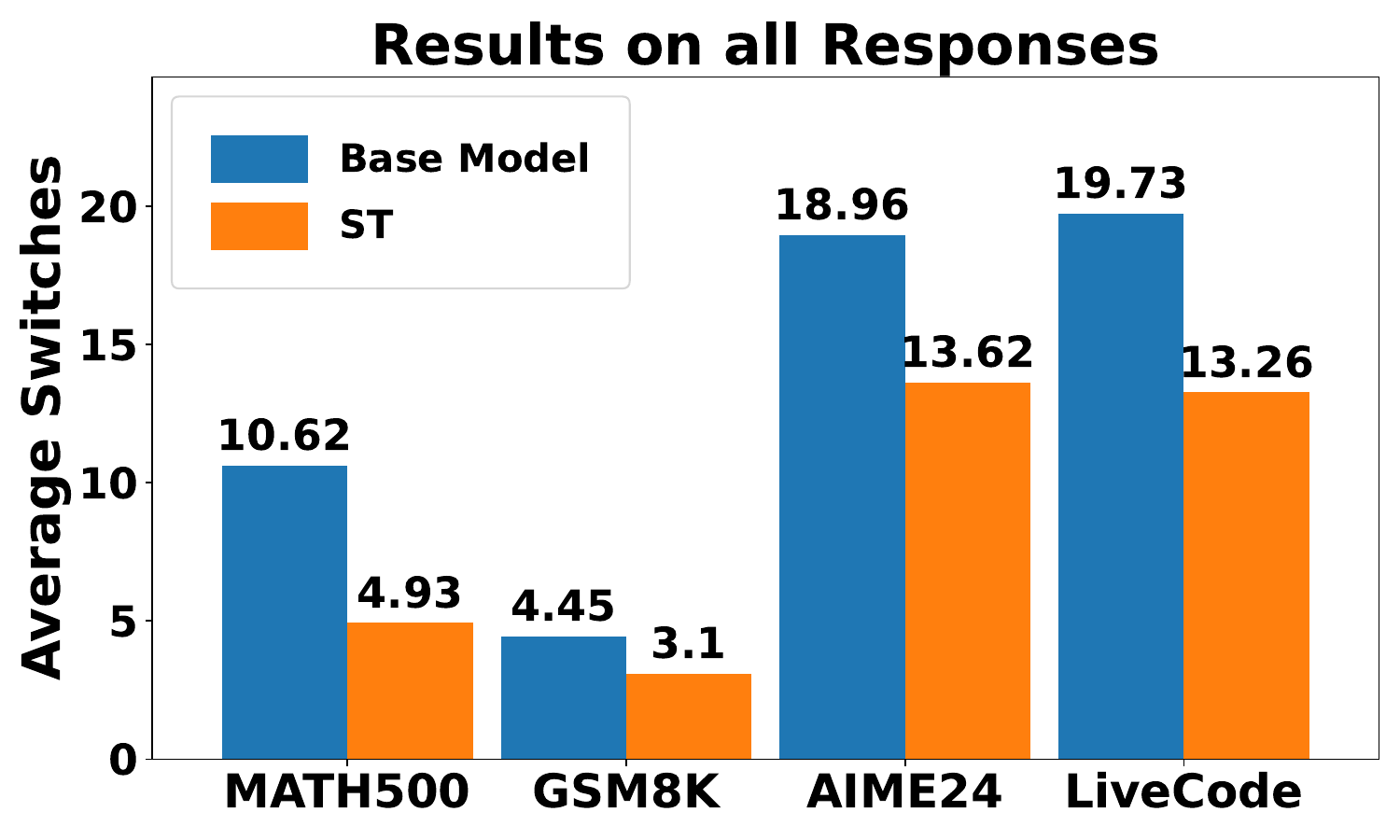}
    \end{subfigure}\hfill
    \begin{subfigure}[b]{0.24\textwidth}
        \centering
        \includegraphics[width=\linewidth]{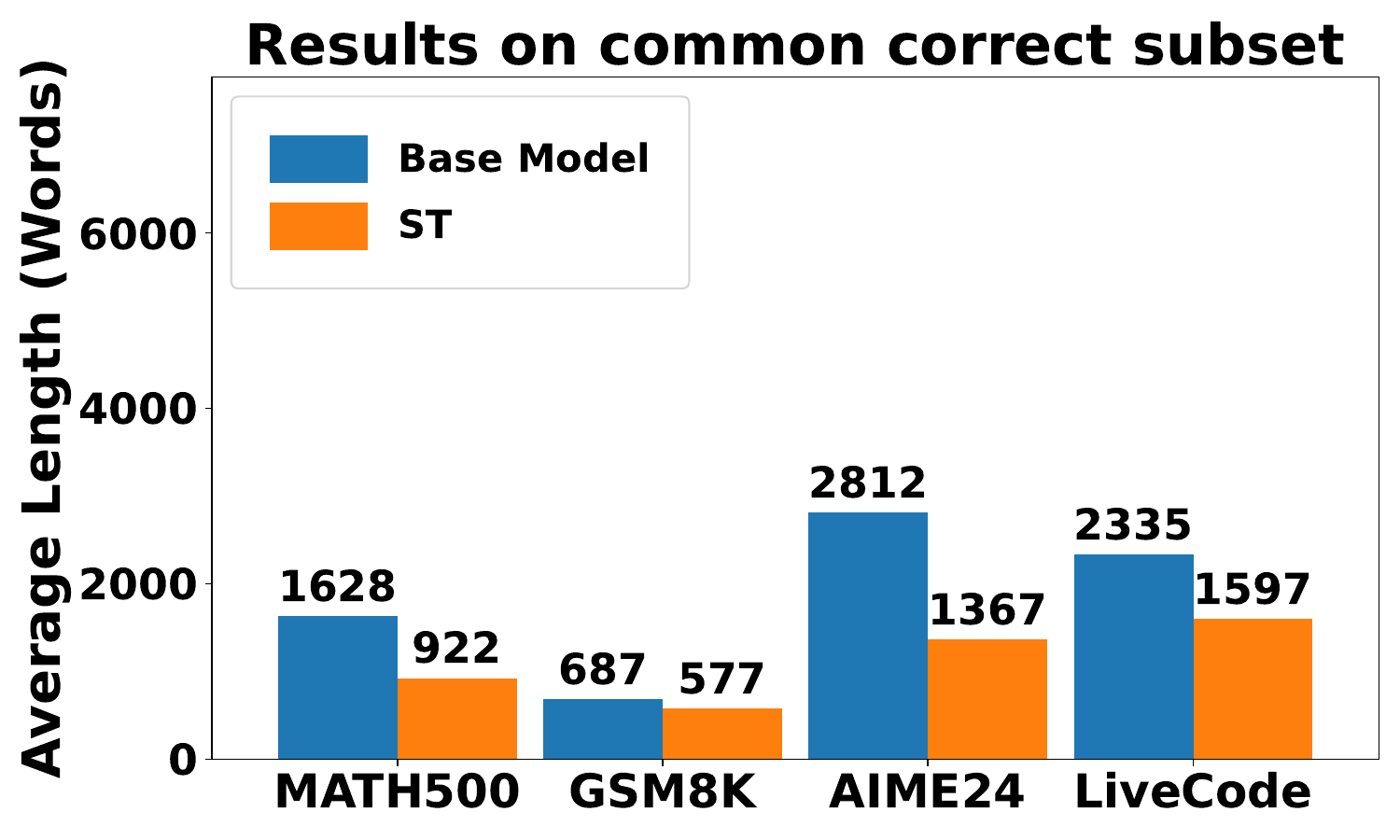}
    \end{subfigure}\hfill
    \begin{subfigure}[b]{0.24\textwidth}
        \centering
        \includegraphics[width=\linewidth]{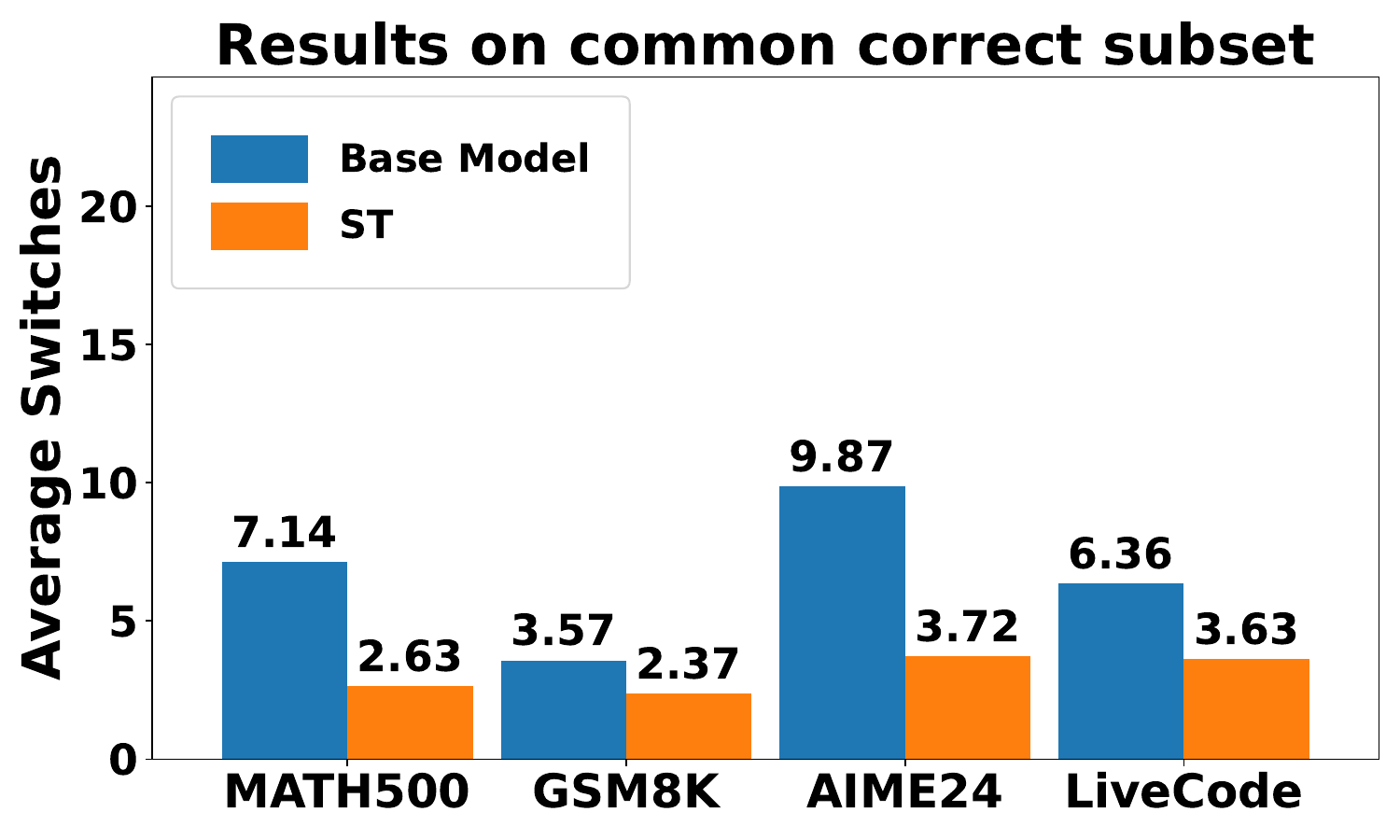}
    \end{subfigure}
    \\[0.1ex] 
    {\small (a) DeepSeek-R1-Distill-Qwen-1.5B} \\[1.2ex] 

    \begin{subfigure}[b]{0.24\textwidth}
        \centering
        \includegraphics[width=\linewidth]{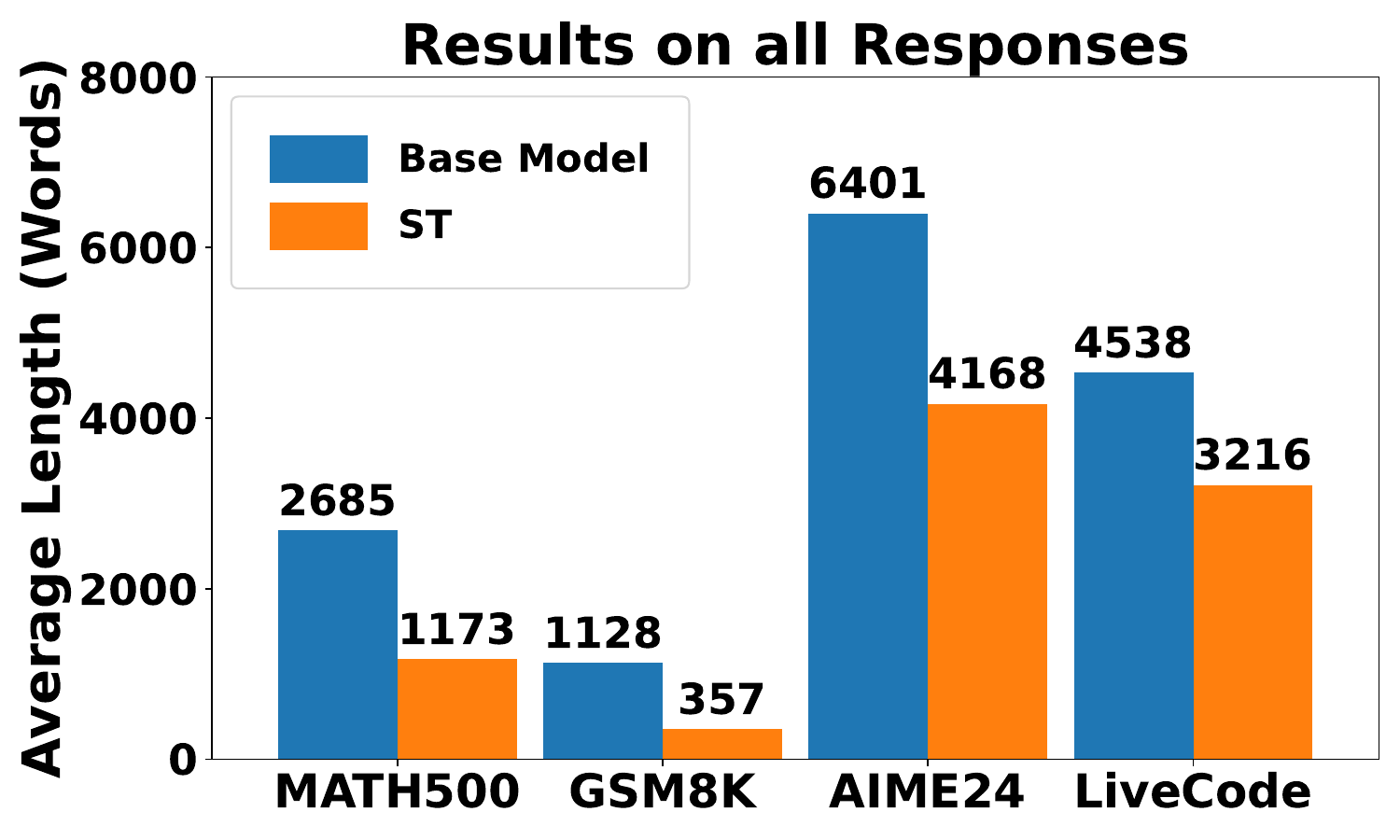}
    \end{subfigure}\hfill
    \begin{subfigure}[b]{0.24\textwidth}
        \centering
        \includegraphics[width=\linewidth]{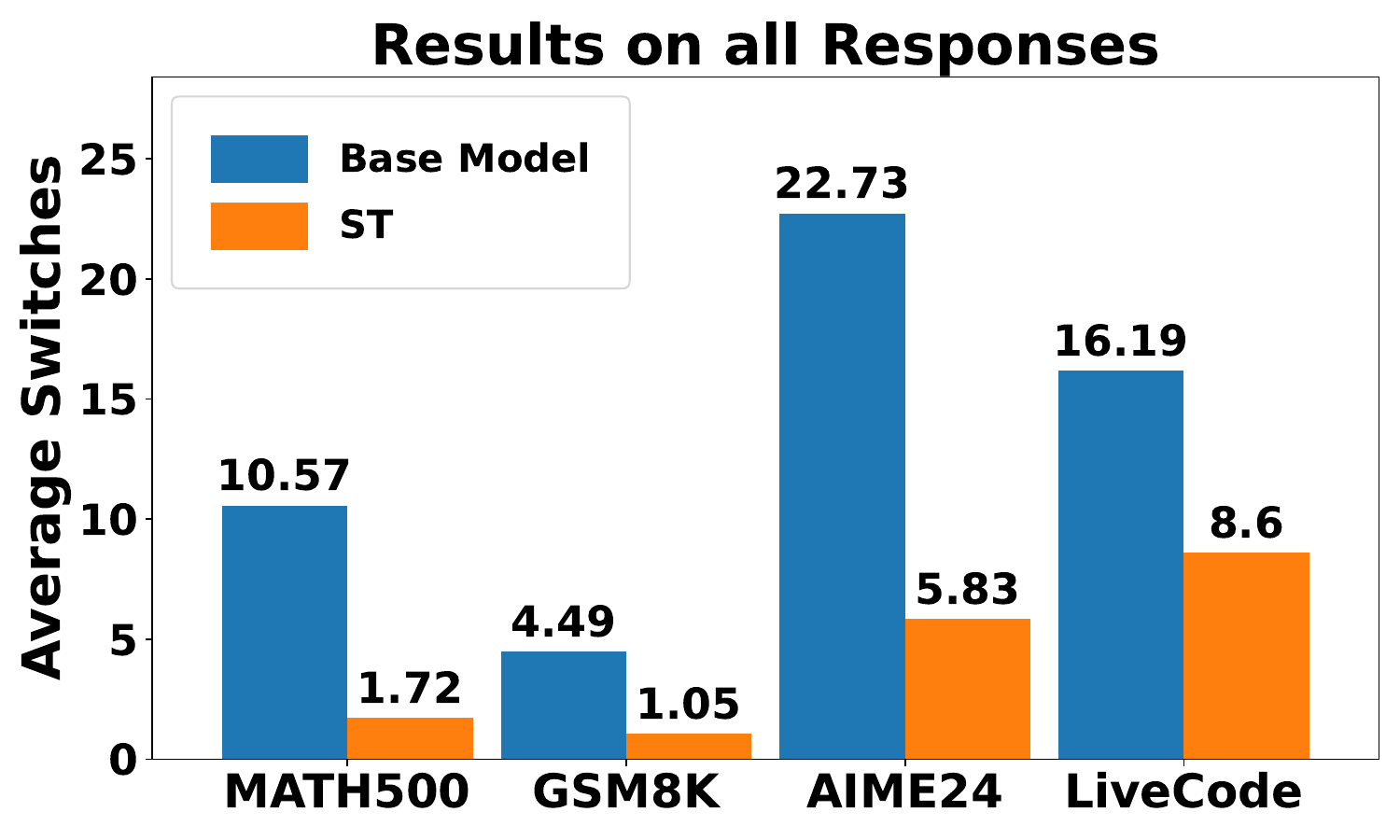}
    \end{subfigure}\hfill
    \begin{subfigure}[b]{0.24\textwidth}
        \centering
        \includegraphics[width=\linewidth]{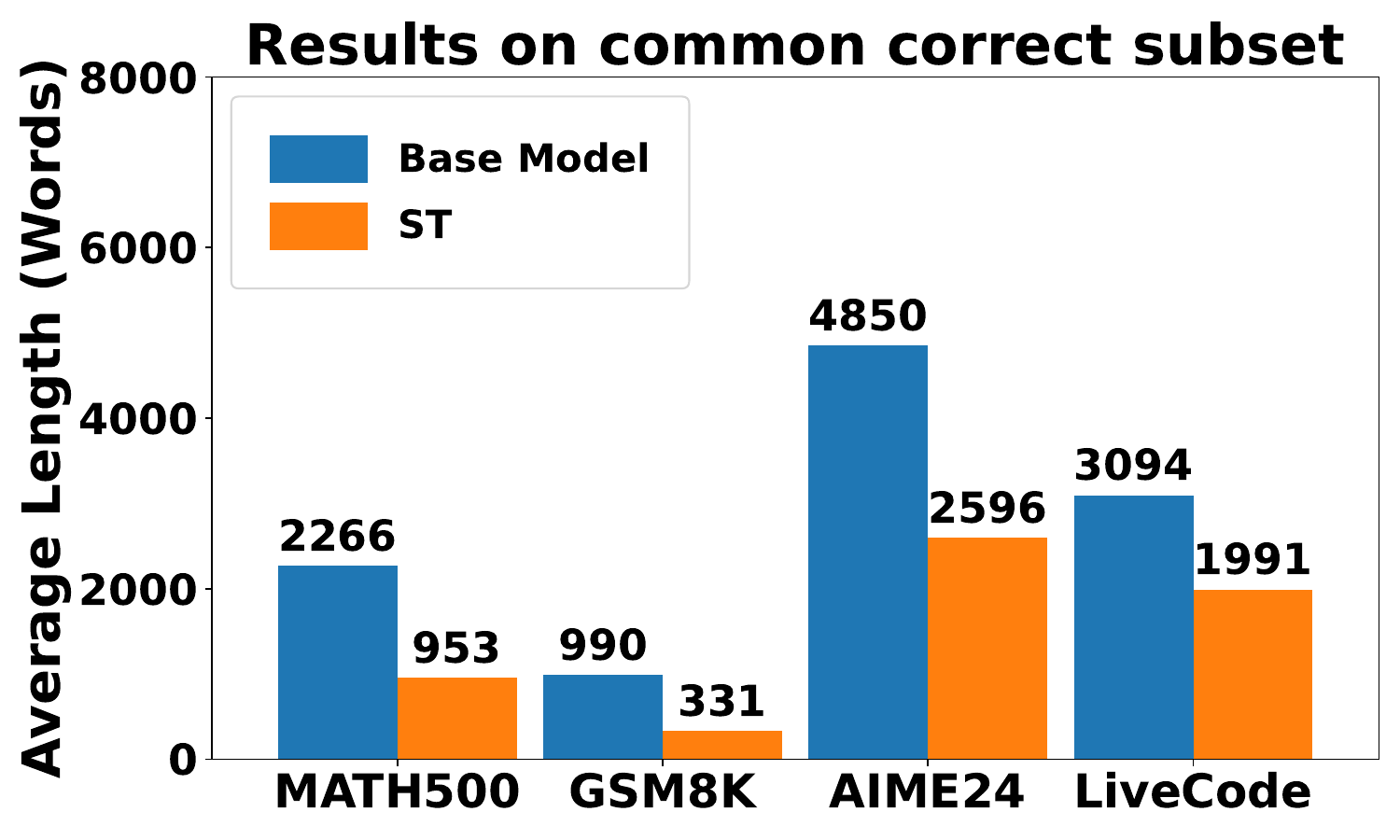}
    \end{subfigure}\hfill
    \begin{subfigure}[b]{0.24\textwidth}
        \centering
        \includegraphics[width=\linewidth]{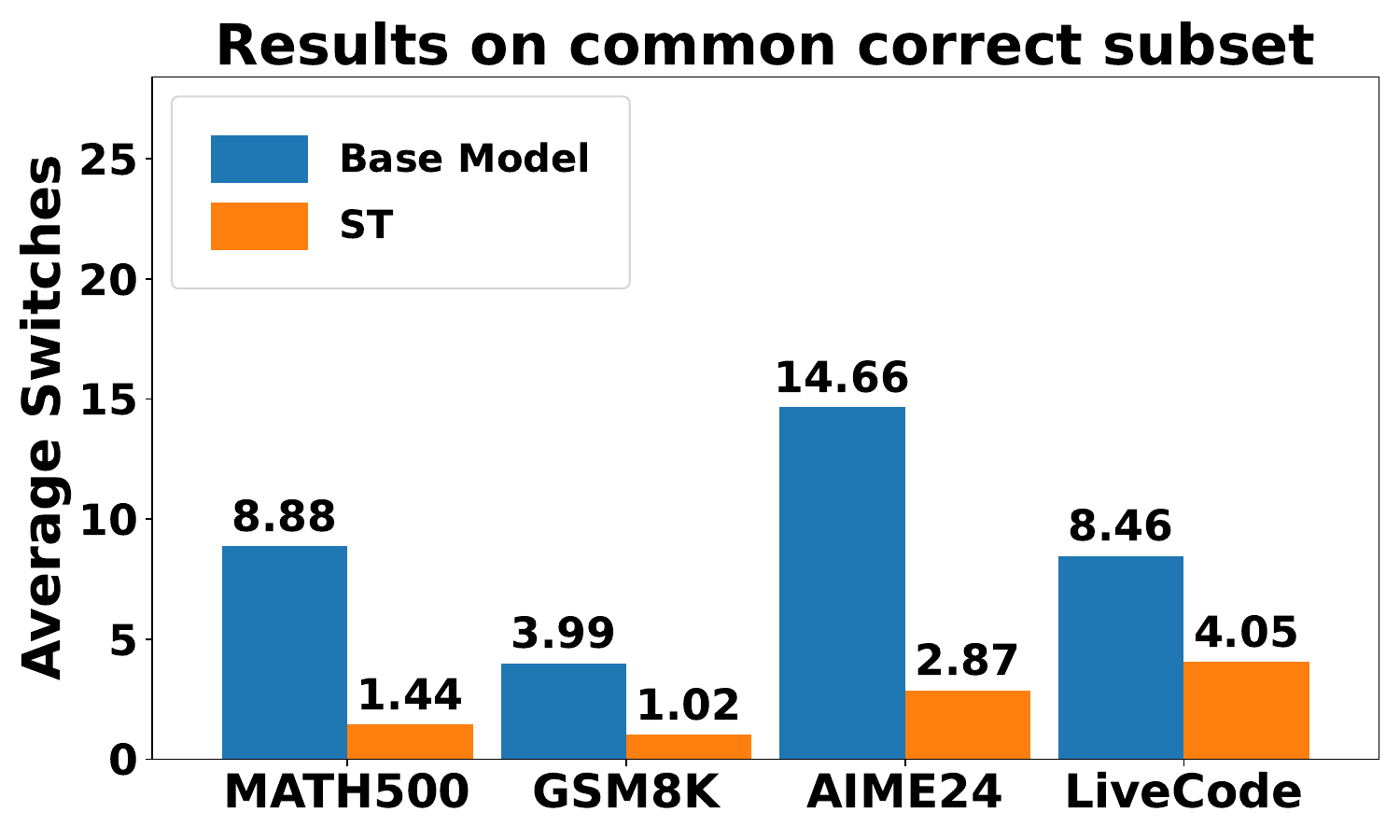}
    \end{subfigure}
    \\[0.1ex]
    {\small (b) Qwen3-8B}

    \caption{Comparison of switching behavior across four metrics: average response length and number of switches on all responses (first two columns) and on the common correct subset (last two columns). Response length is measured in words. The first row shows DeepSeek-R1-Distill-Qwen-1.5B, and the second row shows Qwen3-8B. Blue bars represent the base model, and orange bars represent ST.}
    \label{fig:behavior_analysis-1}
\end{figure*}

ST encourages the model to persist when it reaches a promising thought, thereby reducing unnecessary switches and the resulting redundant generation. This effect is primarily evidenced by the model’s ability to reach correct conclusions with significantly higher efficiency. To evaluate this structural optimization, we conducted a comparative analysis of the model’s behavior before and after ST training across four dimensions: (1) average total output length on all responses, (2) the average number of thought switches on all responses, (3) output length within correct reasoning trajectories, and (4) the average number of switches within those correct trajectories. We measure response length by the number of words.

As illustrated in Figure~\ref{fig:behavior_analysis-1}, models after ST consistently exhibit a reduction in both total output length and the number of thought switches across all datasets. More importantly, this trend persists even for the same set of problems correctly solved both before and after training. By reaching the same correct conclusions with fewer thought-level transitions, the model demonstrates that it has learned to prune redundant reasoning branches and avoid unnecessary switching in its thought process. Consequently, these results provide strong evidence that the observed efficiency gains are driven by more direct reasoning trajectories, in which redundant thought switching is substantially reduced while correct solutions are preserved. We provide further analysis on more models in Appendix~\ref{further switch}.

\begin{table}[t]
\setlength{\textfloatsep}{-10pt}
\centering

\begin{minipage}[t]{0.48\linewidth}
\centering
\caption{Results on the controlled evaluation set.}
\label{tab:recover}
\footnotesize
\renewcommand{\arraystretch}{0.88}
\setlength{\tabcolsep}{10pt}
\begin{tabular}{@{}lcc@{}}
\toprule
\textbf{Method} & \textbf{Acc(\%)}$\uparrow$ & \textbf{Token$\downarrow$} \\
\midrule
\multicolumn{3}{@{}l@{}}{\textit{DeepSeek-R1-Distill-Qwen-1.5B}} \\
\textit{Vanilla} & 13.2 & 7838 \\
\textit{\textbf{Steady Thought}} & 20.3 & 5468 \\
\midrule
\multicolumn{3}{@{}l@{}}{\textit{Qwen3-8B}} \\
\textit{Vanilla} & 23.5 & 9830 \\
\textit{\textbf{Steady Thought}} & 26.6 & 6783 \\
\bottomrule
\end{tabular}
\end{minipage}
\hfill
\begin{minipage}[t]{0.48\linewidth}
\centering
\caption{Results on GAOKAO-bench.}
\label{tab:gaokao}
\footnotesize
\renewcommand{\arraystretch}{0.88}
\setlength{\tabcolsep}{10pt}
\begin{tabular}{@{}lcc@{}}
\toprule
\textbf{Method} & \textbf{Acc(\%)$\uparrow$} & \textbf{Token$\downarrow$} \\
\midrule
\multicolumn{3}{@{}l@{}}{\textit{DeepSeek-R1-Distill-Qwen-1.5B}} \\
\textit{Vanilla} & 80.2 & 2829 \\
\textit{\textbf{Steady Thought}} & 83.7 & 1746 \\
\midrule
\multicolumn{3}{@{}l@{}}{\textit{Qwen3-8B}} \\
\textit{Vanilla} & 86.0 & 4420 \\
\textit{\textbf{Steady Thought}} & 88.4 & 1872 \\
\bottomrule
\end{tabular}
\end{minipage}
\vspace{-10pt}
\end{table}
\subsubsection{Analysis of Recovery Capability}

We further verify that ST does not compromise the model's existing ability to abandon an incorrect reasoning direction when a switch is needed. To this end, we construct a controlled evaluation set from several datasets such as MATH-500 and AIME 2024. We generate responses with the base model and use DeepSeek-V4-Pro to judge the correctness of the first thought, retaining only those with an incorrect initial reasoning direction. The resulting incorrect thoughts are used as fixed prefixes, from which the model continues generation. Token counts are computed only over the generated continuations.

As shown in Table~\ref{tab:recover}, this capability is not impaired by ST. For DeepSeek-R1-Distill-Qwen-1.5B, accuracy increases from 13.2\% to 20.3\%, while the average continuation length decreases from 7,838 to 5,468 tokens. Similarly, for Qwen3-8B, accuracy increases from 23.5\% to 26.6\%, with continuation length decreasing from 9,830 to 6,783 tokens. These results indicate that ST, although reducing thought switching, does not simply enforce persistence in the current reasoning direction when that direction is incorrect. Instead, it preserves the models' ability to recover from an incorrect early thought while generating substantially shorter continuations.

\subsubsection{Analysis of Cross-Lingual Generalization}

To verify that ST does not merely teach the model to reduce the generation of the manually specified switching tokens used during the Trajectory Synthesis stage, we further evaluate its cross-lingual generalization ability. Specifically, we still construct preference data and train the model only on English mathematical datasets, but test it on the Chinese mathematical benchmark GAOKAO-bench~\citep{zhang2023evaluating}. \\
The results are shown in Table~\ref{tab:gaokao}. For DeepSeek-R1-Distill-Qwen-1.5B, ST reduces the average token count by 38.3\% while improving accuracy from 80.2\% to 83.7\%. For Qwen3-8B, ST decreases the average token count by 57.6\% while increasing accuracy from 86.0\% to 88.4\%. These results show that, even when the training and test data differ substantially in language form, ST can still improve accuracy while significantly reducing output length. This suggests that what the model learns through ST is not simply the suppression of certain English switching tokens, but a cross-lingually transferable reasoning behavior pattern.

\subsubsection{Analysis of Different Training Methods}

\setlength{\columnsep}{-1pt}
\begin{wraptable}{r}{0.56\linewidth}
\vspace{-13pt}
\centering
\caption{Comparison of different training methods.}
\label{tab:trainingMethod}
\footnotesize
\renewcommand{\arraystretch}{0.86}
\setlength{\tabcolsep}{0.1pt}

\begin{tabular}{@{}lcccc@{}}
\toprule
\multirow{2}{*}{\textbf{Training Method}} 
& \multicolumn{2}{c}{\textbf{MATH-500}} 
& \multicolumn{2}{c}{\textbf{AIME 2024}} \\
\cmidrule(lr){2-3} \cmidrule(lr){4-5}
& \textbf{Acc(\%)$\uparrow$} & \textbf{Tokens$\downarrow$} 
& \textbf{Acc(\%)$\uparrow$} & \textbf{Tokens$\downarrow$} \\
\midrule
\multicolumn{5}{@{}l}{\textit{\textbf{DeepSeek-R1-Distill-Qwen-1.5B}}} \\
Vanilla & 82.0 & 4385 & 27.5 & 11273 \\
+ SFT-standard & 80.4 & 2650 & 22.9 & 7169 \\
+ SFT-thought & 78.0 & 2661 & 20.8 & 6650 \\
+ DPO-standard & 82.6 & 4273 & 30.8 & 10701 \\
+ DPO-thought & 82.8 & 4428 & 26.7 & 11364 \\
+ SimPO-standard & 81.6 & 4049 & 29.2 & 10592 \\
\textbf{+ STPO} & 83.6 & 2691 & 30.8 & 7577 \\
\midrule
\multicolumn{5}{@{}l}{\textit{\textbf{Qwen3-8B}}} \\
Vanilla & 91.4 & 4724 & 62.1 & 10895 \\
+ SFT-standard & 91.8 & 3181 & 61.7 & 9381 \\
+ SFT-thought & 91.4 & 3020 & 60.8 & 8663 \\
+ DPO-standard & 91.8 & 4495 & 60.8 & 10812 \\
+ DPO-thought & 92.8 & 4594 & 60.0 & 10628 \\
+ SimPO-standard & 93.6 & 2908 & 60.8 & 9151 \\
\textbf{+ STPO} & 93.4 & 2139 & 64.6 & 7640 \\
\bottomrule
\end{tabular}
\vspace{-15pt}
\end{wraptable}

We compare different training strategies in the final stage of ST, as shown in Table~\ref{tab:trainingMethod}. We consider two supervision granularities. In the standard setting, the input is $Q$, with $(T_1,\ldots,T_i,T_i')$ as the chosen target; preference-based methods additionally use the original trajectory $(T_1,\ldots,T_n)$ as the rejected response. In the thought-level setting, the input is $(Q,T_i)$ and the chosen continuation is $T_i'$; DPO-thought and STPO additionally use $(T_{i+1},\ldots,T_n)$ as the rejected continuation. Accordingly, we compare SFT-standard, SFT-thought, DPO-standard, SimPO-standard, DPO-thought, and STPO.

Since SFT learns only from chosen continuations, it mainly encourages imitation of target completions without explicitly distinguishing persistence from unnecessary switching. In contrast, preference optimization directly contrasts these behaviors at the thought-switching point, providing a more targeted signal for sustaining promising thoughts while discouraging unnecessary switches. However, due to our data construction, the chosen continuation is usually much shorter than the rejected one. Since DPO is sensitive to response-length differences, such differences can interfere with preference learning, limiting its effectiveness in reducing unnecessary switching. SimPO alleviates this issue through length-normalized rewards, but SimPO-standard still operates on the entire trajectory and thus provides only coarse-grained supervision. In contrast, STPO combines length normalization with thought-level optimization, reducing the influence of sequence-length signals and focusing the learning objective on the exact divergence point between persistence and unnecessary switching. As a result, STPO reduces token consumption by discouraging unnecessary switching behaviors while maintaining strong reasoning accuracy.

\section{Conclusion}

In this paper, we present Steady Thought (ST), a thought-level preference optimization framework for mitigating under-thinking in Large Reasoning Models. ST identifies thought switching points using complementary linguistic and entropy-based signals, synthesizes persistent continuations, and applies fine-grained preference optimization to discourage unnecessary switching while preserving useful reasoning flexibility. Experiments across multiple models and benchmarks show that ST substantially reduces token consumption while maintaining reasoning accuracy, with token reductions of up to 63.8\%. Its consistent performance on out-of-distribution tasks further suggests that the learned behavior generalizes beyond the training domain. Overall, ST provides an effective approach to improving reasoning efficiency by reducing redundant thought switching.

\section*{Acknowledgements and Disclosures}

We thank the anonymous reviewers for their
valuable comments and constructive suggestions.

\noindent\textbf{Funding.}
This work was supported in part by the New Generation Artificial Intelligence-National Science and Technology Major Project (No. 2025ZD0123102) and the National Natural Science Foundation of China (Grant No. U23A20388).

\noindent\textbf{Competing Interests.}
The authors declare no competing interests.

\bibliography{reference}
\bibliographystyle{plainnat}


\newpage
\appendix
\onecolumn

\section{Related Work}

\subsection{Over-thinking and Under-thinking}

Recent studies have demonstrated strong reasoning capabilities of Large Language Models across diverse tasks
\citep{wang2025logictree,liang2025boosting,liang2026boosting,ma2026styletailor}.
However, these models often consume excessive computational
resources---a phenomenon known as over-thinking \citep{qu2025surveyefficientreasoninglarge,sui2025stopoverthinkingsurveyefficient}. This primarily manifests in two forms: first, the model's responses contain redundant information; second, it struggles to allocate appropriate computational budgets to questions of varying difficulty, with the latter being particularly evident in O1-like models. To address the first issue, existing solutions include both training-free and training methods. Training-free methods involve carefully designed prompts to limit output length \citep{xu2025chaindraftthinkingfaster,nayab2025concisethoughtsimpactoutput} or adjusting the model's output during decoding to enhance reasoning conciseness \citep{qiu2024treebonenhancinginferencetimealignment,zhang2025lightthinkerthinkingstepbystepcompression}. Training methods improve the model's ability to provide concise answers by fine-tuning on concise chain-of-thought (CoT) data or incorporating additional rewards/penalties for output length during reinforcement learning (RL) training \citep{kang2024c3otgeneratingshorterchainofthought,aggarwal2025l1controllinglongreasoning}. For the second issue, most approaches employ fine-tuning or RL to enable models to dynamically switch between fast-thinking and slow-thinking modes during reasoning based on question difficulty \citep{zhang2025adaptthinkreasoningmodelslearn,lou2025adacotparetooptimaladaptivechainofthought} or leverage latent guidance \citep{chen2026latent}, thereby improving the balance between reasoning accuracy and efficiency . Under-thinking represents another form of resource wastage, primarily caused by LRMs frequently switching between reasoning directions before fully developing promising thoughts \citep{wang2025thoughtsplaceunderthinkingo1like}. Existing methods focus on directly suppressing the model's thought switching behavior at either the token or representation level to mitigate this phenomenon \citep{ding2025thinkingtokenshelptrap,chen2025sealsteerablereasoningcalibration,wang2025waitdontneedwait}. In contrast, our approach considers that some switching is necessary, and instead alleviates under-thinking by enhancing the model's ability to persist in promising reasoning thoughts.

\subsection{Preference Optimization}
Preference optimization aims to align model behavior with human or task-specific preferences~\citep{liang2025ropo}. For example, some Proximal Policy Optimization (PPO)-based methods typically train a reward model from human preference data and then optimize the policy against this reward while constraining the magnitude of policy updates~\citep{lightmanlet,luo2023wizardmath}. However, such pipelines are typically complex, requiring multiple training stages and additional models. Direct Preference Optimization (DPO) ~\citep{rafailov2024direct} method simplifies this process by directly optimizing the policy from preference pairs without explicitly training a reward model. Subsequent studies have extended DPO to finer levels of supervision, including step-level and token-level optimization ~\citep{lu2024stepcontrolled,lai2024stepdpo,liu2024tisdpo}. Nevertheless, DPO can be sensitive to substantial length differences between preferred and rejected responses because its preference objective is not explicitly normalized by response length. In contrast, Simple Preference Optimization (SimPO) ~\citep{meng2024simposimplepreferenceoptimization} addresses this issue by using average log-probability as a length-normalized implicit reward, thereby reducing the influence of response length on preference learning. Inspired by SimPO, we further introduce a finer-grained preference optimization framework, enabling the model to learn thought-level responses to mitigate the under-thinking issue.

\section{Implementation Details}
\subsection{Training and Inference Details}
For all models, we use a batch size of 256. For the 1.5B model, we use a learning rate of $5\times10^{-7}$, $\gamma = 1.0$, and $\beta = 2.0$. The 8B, 14B, and 32B models are trained using Low-Rank Adaptation (LoRA). Specifically, the 8B and 32B models use a learning rate of $3\times10^{-5}$ and $\beta = 2.0$, while the 14B model uses a learning rate of $13\times10^{-5}$ and $\beta = 3.0$. The entropy thresholds $\tau$ for the 1.5B, 8B, 14B, and 32B models are set to 2.2, 1.3, 1.9, and 1.3, respectively.

Across all experiments, models were optimized using Adam with a context window of 6144. Our training framework incorporated DeepSpeed ZeRO-3 with CPU offload for memory optimization, and we implemented a warm-up ratio of 0.1 with a cosine learning rate scheduler. For all inference tasks, we maintained uniform generation parameters: $\mathit{max\_tokens}=14336$, $\mathit{temperature}=0.6$, and $\mathit{top\_p}=0.95$.

\subsection{Explicit Linguistic Markers}
\label{Markers}

Table~\ref{tab:phrase} lists the predefined linguistic markers used to identify explicit thought switching signals. These phrases typically indicate that the model is reconsidering its current reasoning direction, introducing an alternative approach, or transitioning to a different strategy. During switching-point detection, the linguistic-marker condition is satisfied when the subsequent step begins with one of these predefined phrases.

\begin{table}[htbp]
\centering
\caption{Explicit linguistic markers}
\label{tab:phrase}
\small
\begin{tabular}{@{}lll@{}} 
\toprule
\texttt{alternatively}     &  \texttt{another way}      & \texttt{another thought}    \\
\texttt{another approach}  & \texttt{another method}    &  \texttt{another solution}     \\
\texttt{another technique}           & \texttt{think differently}    & \texttt{another strategy}          \\ 
\bottomrule
\end{tabular}
\end{table}

\subsection{Token Set for Forced Continuation}
During the trajectory synthesis stage of ST, we encourage the model to continue along the current reasoning direction by applying a large negative logit bias to tokens that commonly signal a transition to a new thought. The predefined token set used for forced continuation is listed in Table~\ref{tab:tokens}.

\begin{table}[htbp]
\caption{Token set used for forced continuation during trajectory synthesis.}

\centering
\small
\adjustbox{max width=\textwidth}{%

\begin{tabular}{lllllllll}
\toprule
\texttt{(any} & \texttt{-any} & \texttt{.Any} & \texttt{.any} & \texttt{:Any} & \texttt{:any} & \texttt{<Any} & \texttt{<any} & \texttt{Any} \\
\texttt{[Any} & \texttt{\_any} & \texttt{any} & \texttt{ĠAny} & \texttt{(other} & \texttt{-other} & \texttt{.Other} & \texttt{.other} & \texttt{Other} \\
\texttt{\_other} & \texttt{other} & \texttt{ĠOther} & \texttt{Ġother} & \texttt{Hmm} & \texttt{ĠHmm} & \texttt{Ġhmm} & \texttt{Alternative} & \texttt{alternative} \\
\texttt{ĠAlternative} & \texttt{Ġalternative} & \texttt{Another} & \texttt{another} & \texttt{ĠAnother} & \texttt{Ġanother} & \texttt{(check} & \texttt{-check} & \texttt{.Check} \\
\texttt{.check} & \texttt{/check} & \texttt{Check} & \texttt{\_Check} & \texttt{\_check} & \texttt{check} & \texttt{ĉCheck} & \texttt{ĉcheck} & \texttt{ĠCheck} \\
\texttt{Ġcheck} & \texttt{Alternatively} & \texttt{ĠAlternatively} & \texttt{Ġalternatively} & \texttt{"But} & \texttt{,but} & \texttt{-but} & \texttt{.But} & \texttt{.but} \\
\texttt{But} & \texttt{\_but} & \texttt{but} & \texttt{âĢ¦but} & \texttt{âĢĶbut} & \texttt{âĢľBut} & \texttt{ĠBut} & \texttt{Ġbut} & \texttt{However} \\
\texttt{however} & \texttt{ĠHowever} & \texttt{Ġhowever} & \texttt{.Verify} & \texttt{.verify} & \texttt{Verify} & \texttt{\_verify} & \texttt{verify} & \texttt{ĉverify} \\
\texttt{ĠVerify} & \texttt{Ġverify} & \texttt{Again} & \texttt{\_again} & \texttt{again} & \texttt{ĠAgain} & \texttt{Ġagain} & \texttt{(wait} & \texttt{.Wait} \\
\texttt{.wait} & \texttt{/wait} & \texttt{Wait} & \texttt{\_wait} & \texttt{wait} & \texttt{ĉwait} & \texttt{ĠWait} & \texttt{Ġwait} & \texttt{"Now} \\
\texttt{(now} & \texttt{-now} & \texttt{.Now} & \texttt{.now} & \texttt{Now} & \texttt{[now} & \texttt{\_now} & \texttt{now} & \texttt{ĉnow} \\
\texttt{ĠNow} & \texttt{Ġnow} & \texttt{Ah} & \texttt{ah} & \texttt{aha} & \texttt{ĠAh} & \texttt{Ġah} & \texttt{"Oh} & \texttt{Oh} \\
\texttt{oh} & \texttt{âĢľOh} & \texttt{ĠOh} & \texttt{Ġoh} & \texttt{.Maybe} & \texttt{Maybe} & \texttt{maybe} & \texttt{ĠMaybe} & \texttt{Ġmaybe} \\
\bottomrule
\end{tabular}%
}

\label{tab:tokens}
\end{table}

\section{Experiments on Different Model Family}

\begin{table*}[t]
\centering
\caption{Results on DeepSeek-R1-Distill-Llama-8B across different benchmarks.}
\label{tab:llama8b}
\footnotesize
\renewcommand{\arraystretch}{0.86}
\setlength{\tabcolsep}{5.0pt}
\begin{tabular}{@{}lcccccccc@{}}
\toprule
\multirow{2}{*}{\textbf{Method}} 
& \multicolumn{2}{c}{\textbf{MATH-500}} 
& \multicolumn{2}{c}{\textbf{AIME 2024}}
& \multicolumn{2}{c}{\textbf{GSM8K}}
& \multicolumn{2}{c}{\textbf{LiveCode}} \\
\cmidrule(lr){2-3} 
\cmidrule(lr){4-5}
\cmidrule(lr){6-7}
\cmidrule(lr){8-9}
& \textbf{Acc(\%)$\uparrow$} & \textbf{Tokens$\downarrow$} 
& \textbf{Acc(\%)$\uparrow$} & \textbf{Tokens$\downarrow$}
& \textbf{Acc(\%)$\uparrow$} & \textbf{Tokens$\downarrow$}
& \textbf{Acc(\%)$\uparrow$} & \textbf{Tokens$\downarrow$} \\
\midrule
Vanilla 
& 82.8 & 3369 
& 41.7 & 9638 
& \textbf{88.1} & 1112 
& 56.4 & 8398 \\

\textbf{Steady Thought} 
& 85.8 & 2653
& 44.2 & 9100
& 87.4 & 874
& 56.8 & 7447 \\
\bottomrule
\end{tabular}
\end{table*}

To further examine whether the effectiveness of Steady Thought generalizes beyond the Qwen model family, we evaluate ST on DeepSeek-R1-Distill-Llama-8B across multiple reasoning benchmarks. As shown in Table~\ref{tab:llama8b}, ST consistently reduces token consumption across all evaluated benchmarks. It also improves accuracy on most datasets. These results suggest that the effectiveness of Steady Thought is not limited to Qwen-based reasoning models and can transfer to a different model family.

\section{Further Analysis of Switching Ability}
\label{further switch}
\begin{figure*}[]

    \centering
    
    \begin{subfigure}[b]{0.24\textwidth}
        \centering
        \includegraphics[width=\linewidth]{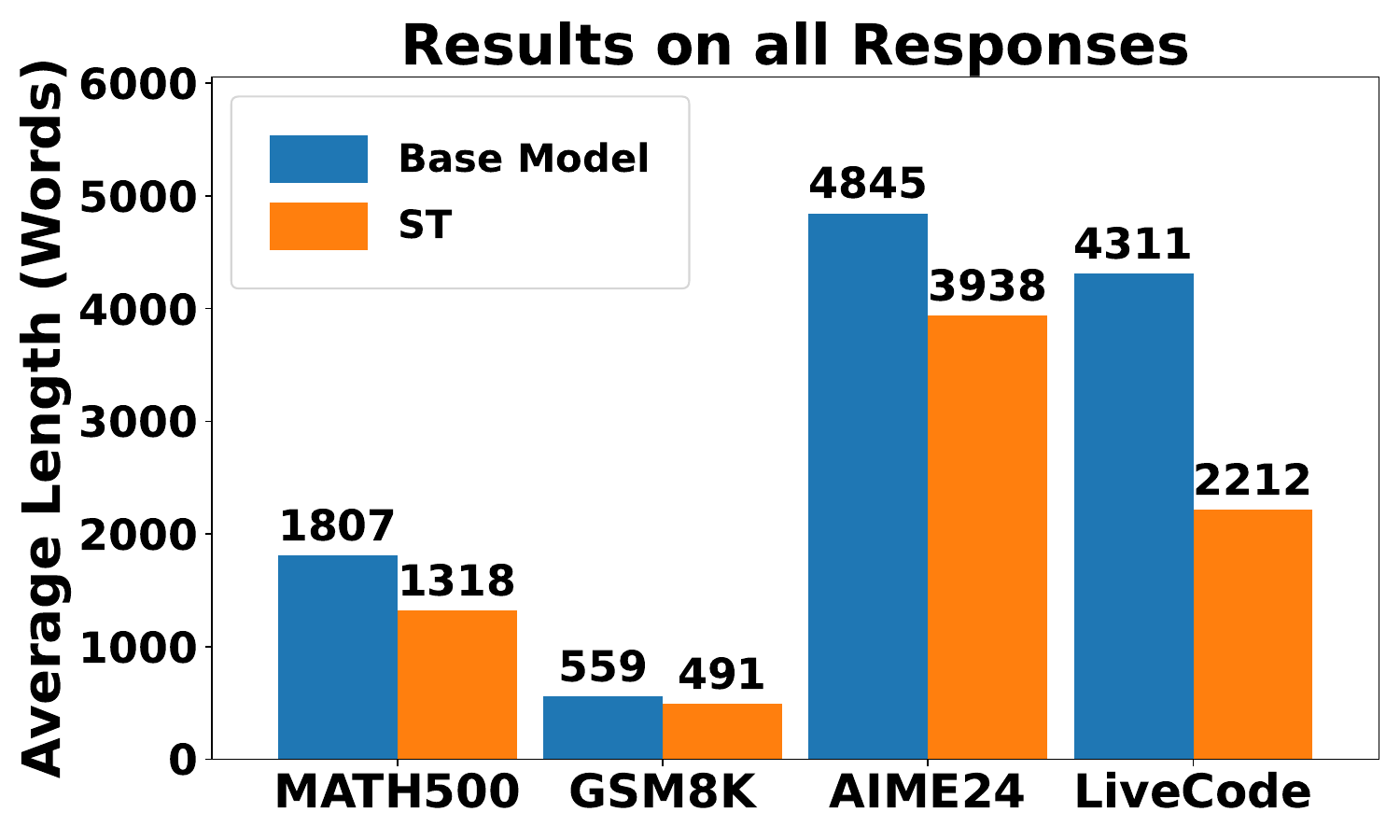}
    \end{subfigure}\hfill
    \begin{subfigure}[b]{0.24\textwidth}
        \centering
        \includegraphics[width=\linewidth]{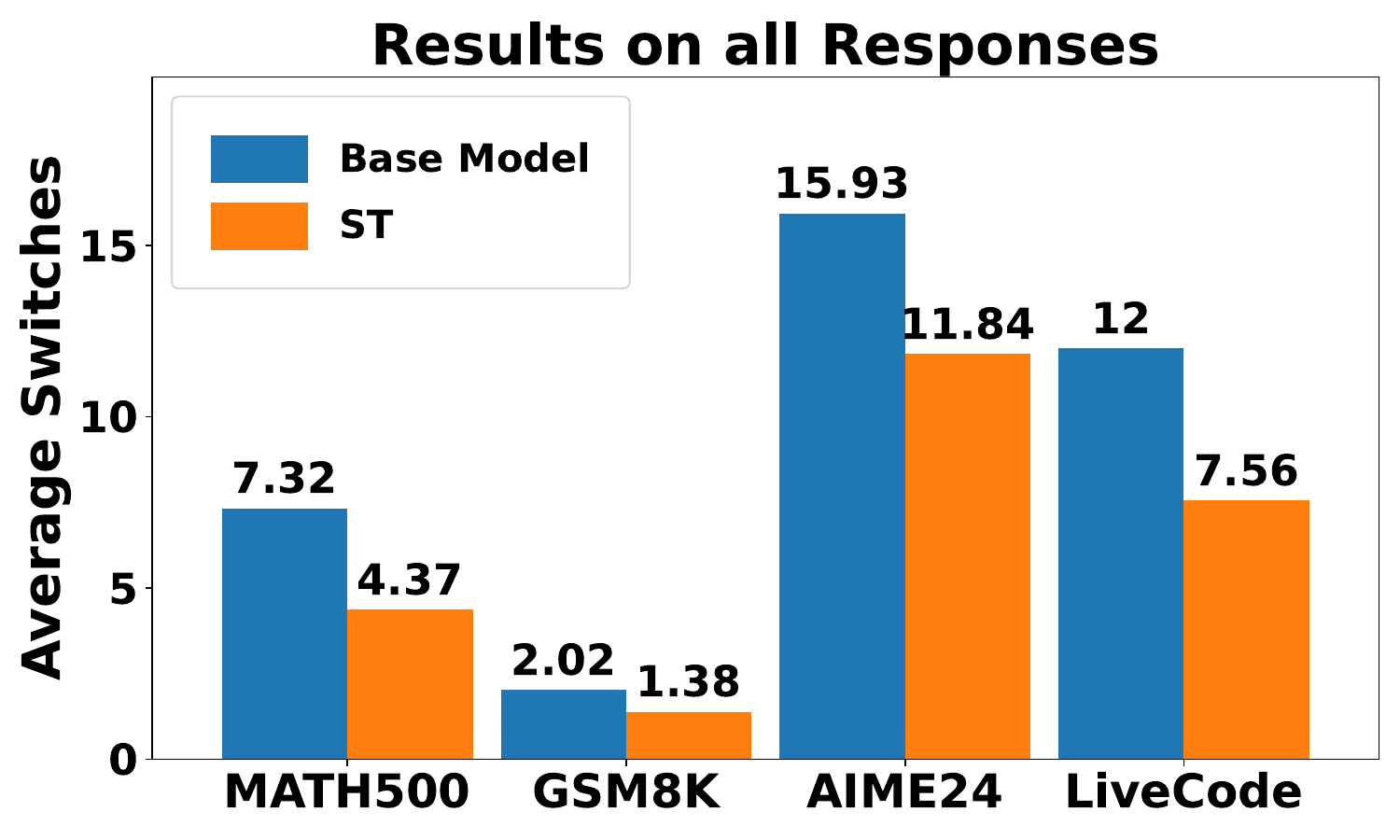}
    \end{subfigure}\hfill
    \begin{subfigure}[b]{0.24\textwidth}
        \centering
        \includegraphics[width=\linewidth]{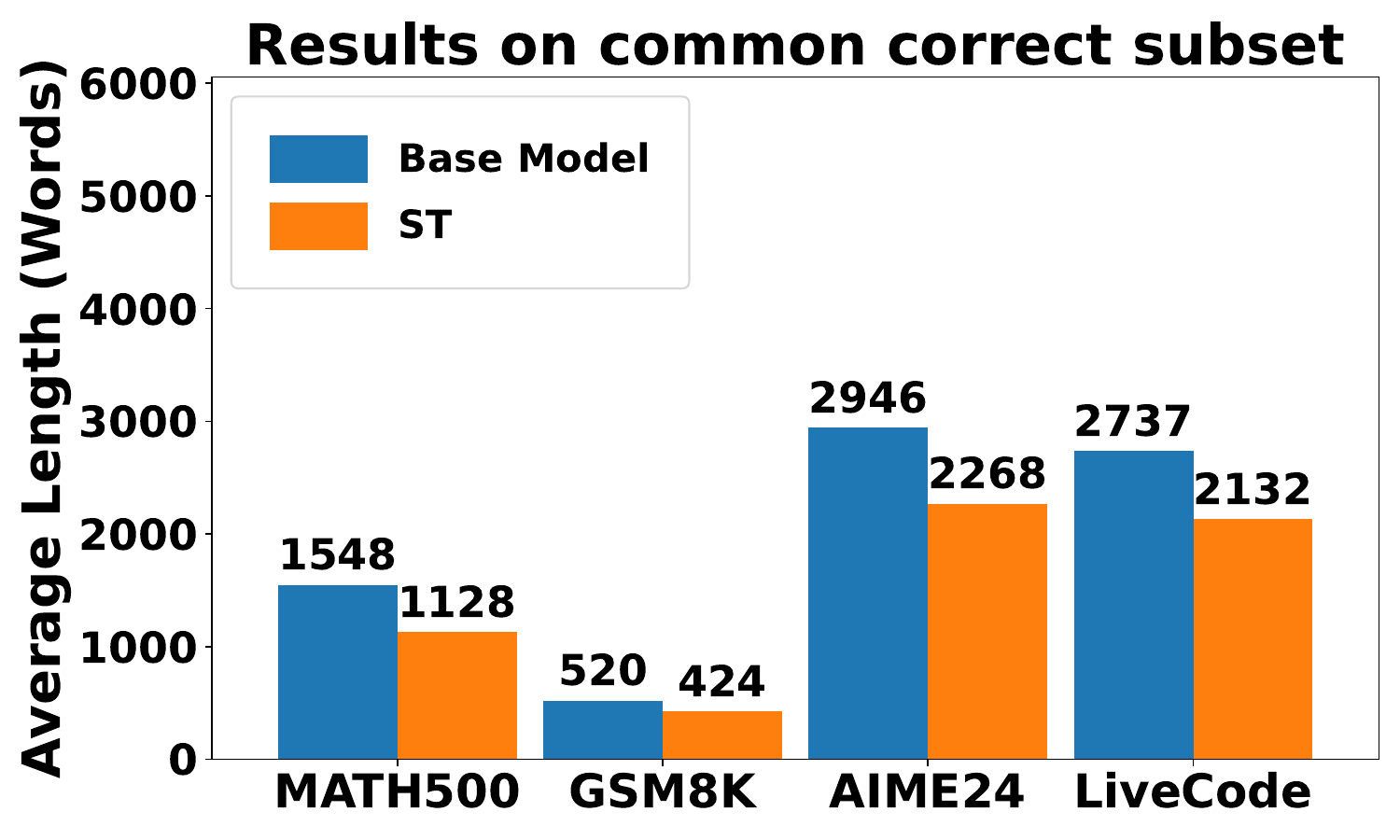}
    \end{subfigure}\hfill
    \begin{subfigure}[b]{0.24\textwidth}
        \centering
        \includegraphics[width=\linewidth]{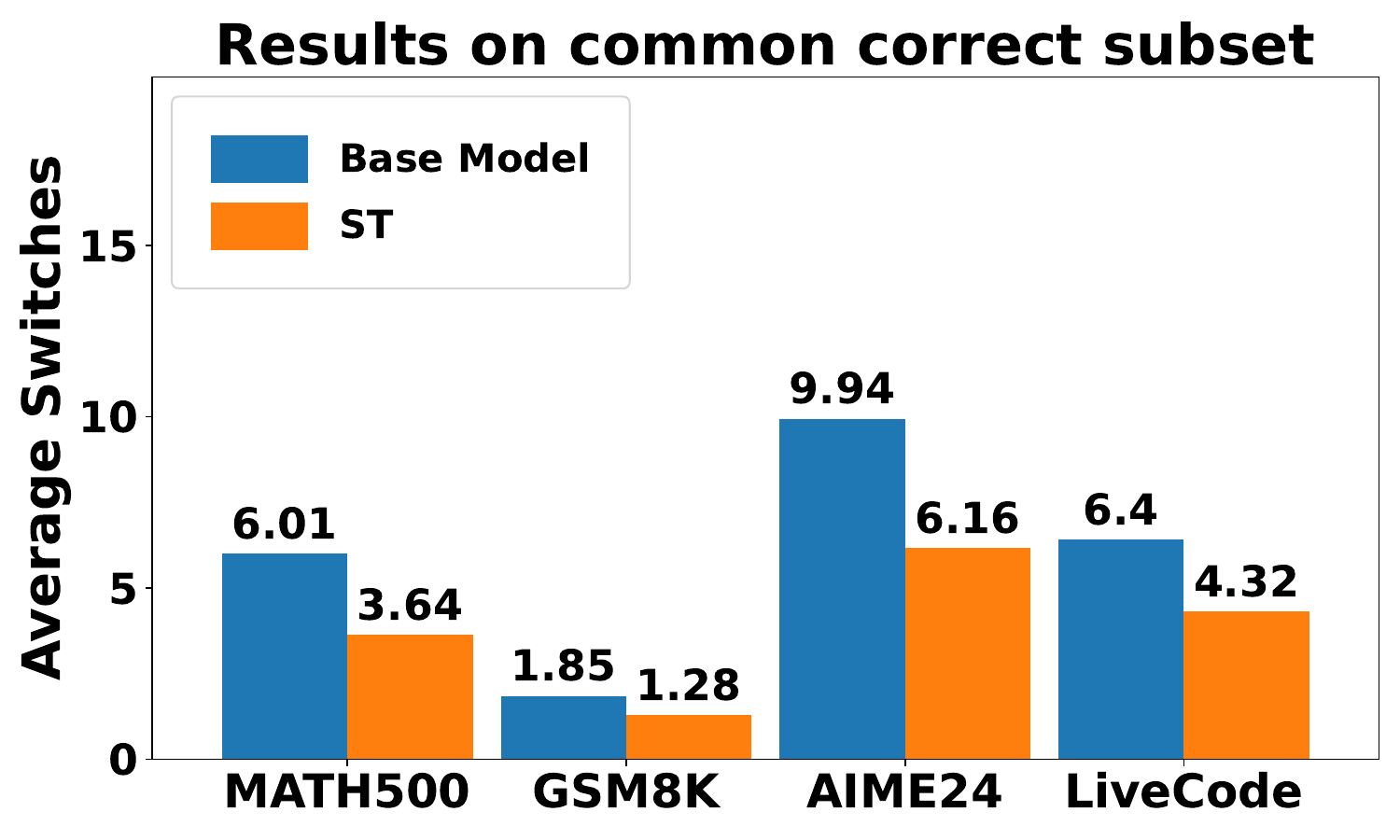}
    \end{subfigure}
    \\[0.5ex] 
    {\small (a) DeepSeek-R1-Distill-Qwen-14B} \\[1.2ex] 

    \begin{subfigure}[b]{0.24\textwidth}
        \centering
        \includegraphics[width=\linewidth]{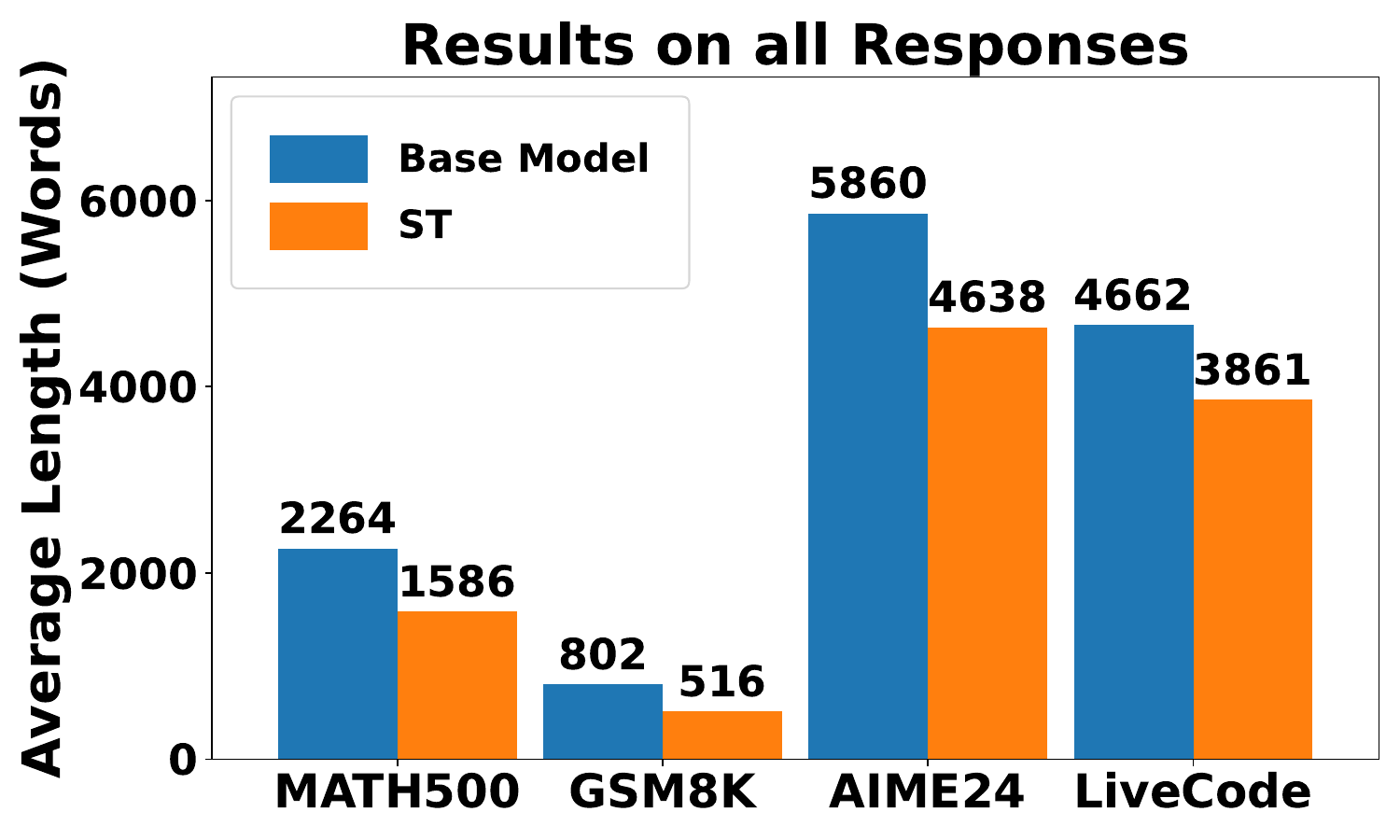}
    \end{subfigure}\hfill
    \begin{subfigure}[b]{0.24\textwidth}
        \centering
        \includegraphics[width=\linewidth]{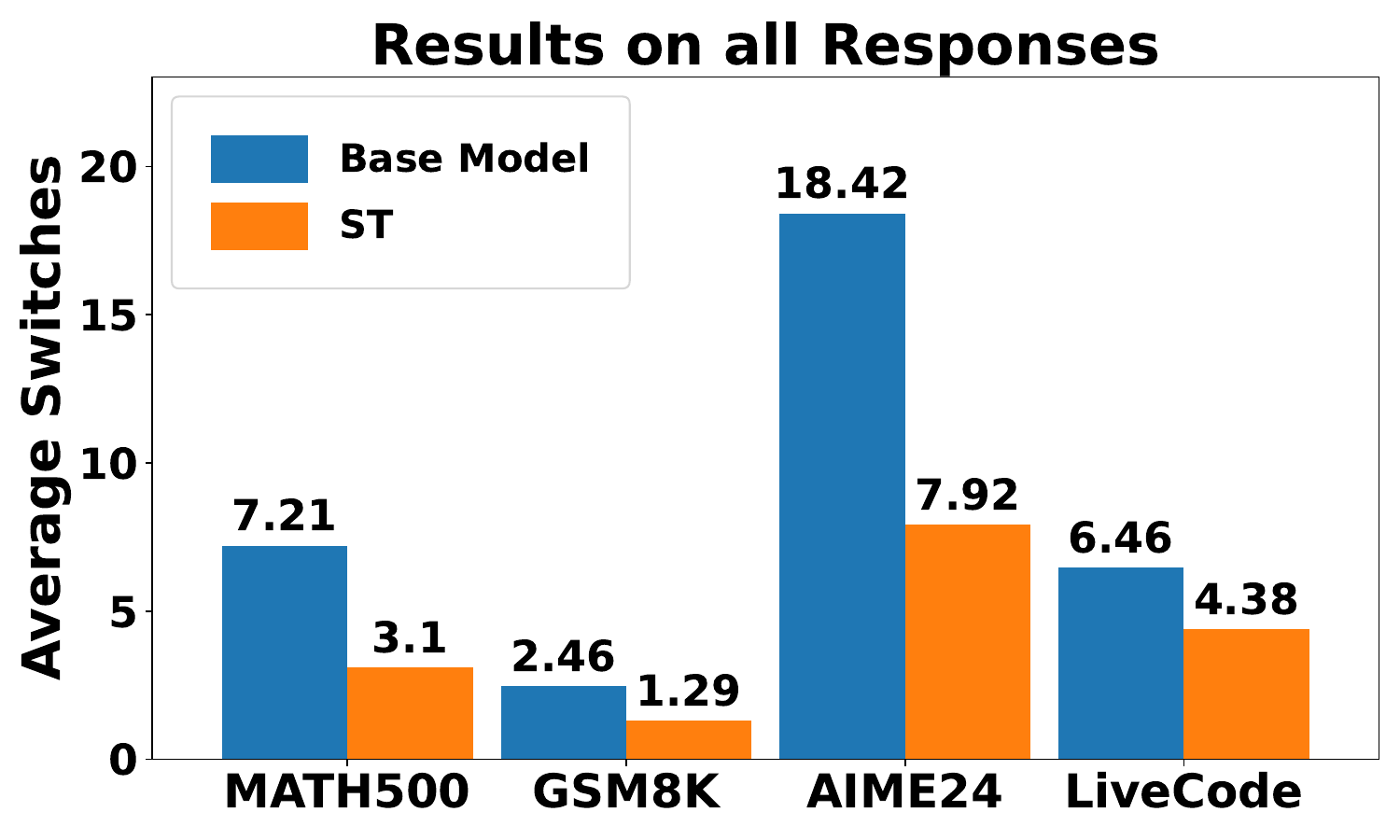}
    \end{subfigure}\hfill
    \begin{subfigure}[b]{0.24\textwidth}
        \centering
        \includegraphics[width=\linewidth]{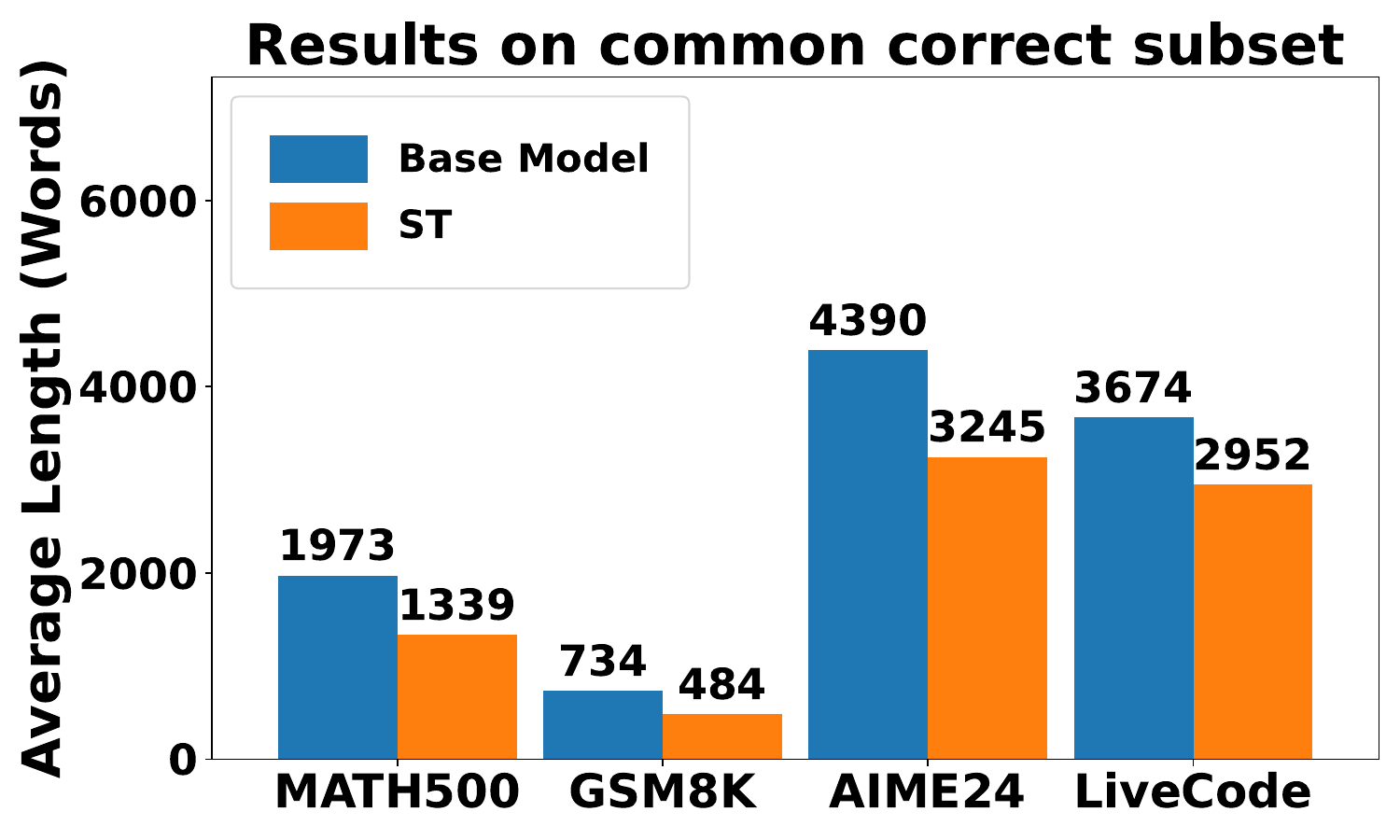}
    \end{subfigure}\hfill
    \begin{subfigure}[b]{0.24\textwidth}
        \centering
        \includegraphics[width=\linewidth]{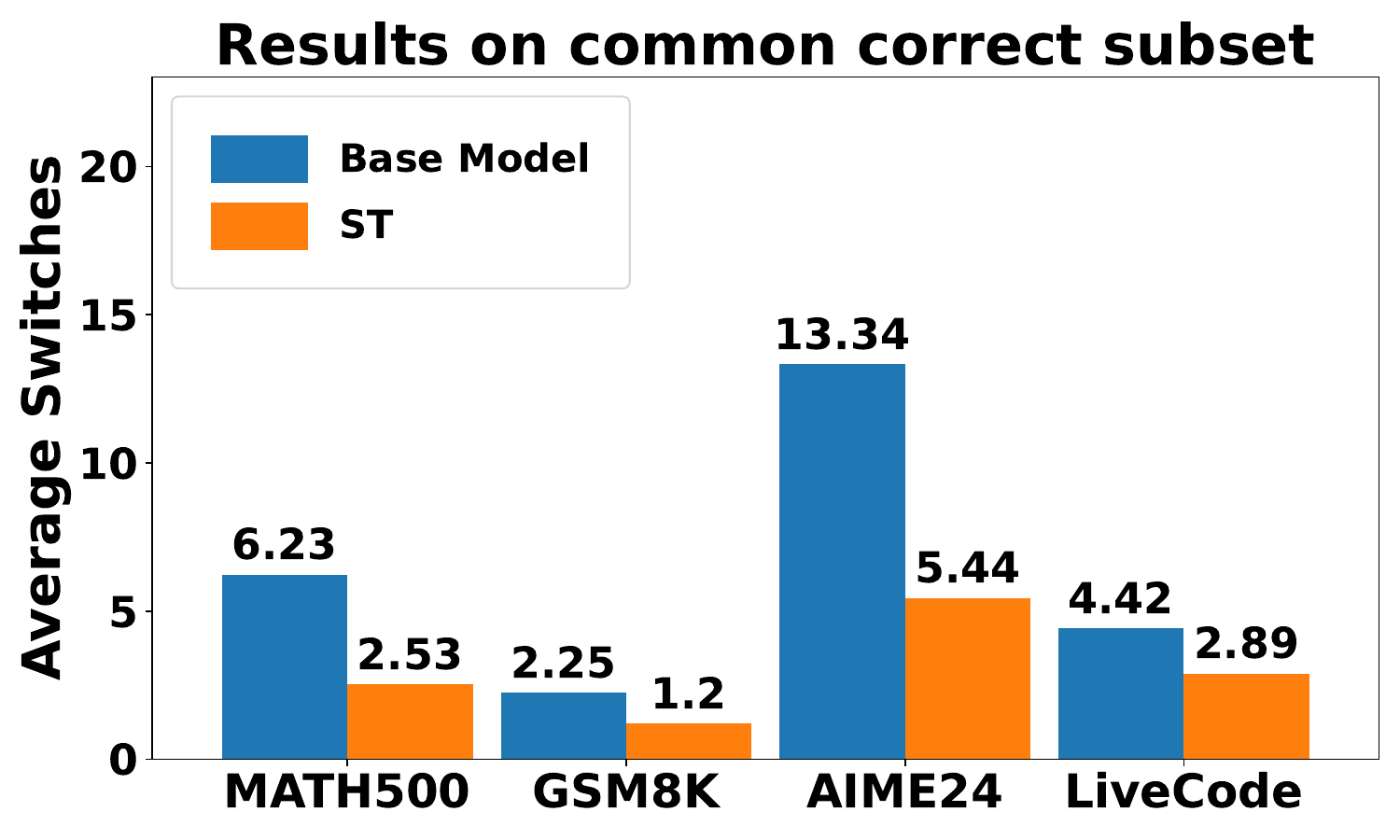}
    \end{subfigure}
    \\[0.5ex]
    {\small (b) Qwen3-32B}

    \vspace{0.5em} 
    \caption{Comparison of switching behavior across four metrics: average response length and number of switches on all responses (first two columns) and on the common correct subset (last two columns). Response length is measured in words. The first row shows DeepSeek-R1-Distill-Qwen-14B, and the second row shows Qwen3-32B. Blue bars represent the base model, and orange bars represent ST.}
    \label{fig:behavior_analysis}
\end{figure*}

We analyze the variations in thought switching capabilities for the 14B and 32B models before and after the application of ST. As shown in Figure~\ref{fig:behavior_analysis}, consistent with the results observed in small-scale models, these larger-parameter models demonstrate significantly reduced response length for problems that were already correctly solved by the base models following ST training.

\section{Ablation on Switching-Point Signals}
\label{Switching-Point}
To isolate the contributions of the two signals used for identifying thought switching points, we conduct an ablation study on explicit linguistic markers and boundary entropy spikes. The results are reported in Table~\ref{tab:signal_ablation}. The full ST method achieves the best overall trade-off between accuracy and token efficiency, suggesting that the two signals are complementary and that combining them is more effective than relying on either signal alone.

\begin{table}[t]
\centering
\caption{Ablation study on switching-point signals.}
\label{tab:signal_ablation}
\renewcommand{\arraystretch}{0.95}
\setlength{\tabcolsep}{5pt}
\begin{tabular}{lcccc}
\toprule
\multirow{2}{*}{\textbf{Method}} 
& \multicolumn{2}{c}{\textbf{MATH-500}} 
& \multicolumn{2}{c}{\textbf{AIME 2024}} \\
\cmidrule(lr){2-3} \cmidrule(lr){4-5}
& \textbf{Acc(\%)$\uparrow$} & \textbf{Tokens$\downarrow$} 
& \textbf{Acc(\%)$\uparrow$} & \textbf{Tokens$\downarrow$} \\
\midrule
DeepSeek-R1-Distill-Qwen-1.5B & 82.0 & 4385 & 27.5 & 11273 \\
\textbf{+ Steady Thought} & \textbf{83.6} & 2691 & \textbf{30.8} & \textbf{7577} \\
\quad w/o explicit linguistic markers & 82.2 & \textbf{2609} & 29.2 & 8316 \\
\quad w/o boundary entropy spikes & 83.4 & 3554 & 28.7 & 8199 \\
\midrule
Qwen3-8B & 91.4 & 4724 & 62.1 & 10895 \\
\textbf{+ Steady Thought} & \textbf{93.4} & \textbf{2139} & \textbf{64.6} & \textbf{7640} \\
\quad w/o explicit linguistic markers & \textbf{93.4} & 3528 & 63.3 & 9712 \\
\quad w/o boundary entropy spikes & 93.0 & 3642 & 62.9 & 9685 \\
\bottomrule
\end{tabular}
\end{table}

\section{Analysis of Different Entropy Thresholds}
\label{thresholds}

For threshold calibration, we sample 60 responses per model from the training pool and evaluate candidate thresholds in ascending order based solely on segmentation quality. Specifically, we assess whether the detected boundaries correspond to genuine switches in reasoning direction, using human judgment for clear cases and an LLM judge only for ambiguous ones. We select the lowest candidate threshold for which at least 80\% of the detected boundaries are judged to be genuine thought switches. Since increasing $\tau$ progressively reduces the number of detected boundaries, this criterion preserves as much boundary coverage as possible while maintaining reliable segmentation.

We investigate the sensitivity of Steady Thought to the entropy threshold $\tau$ used for identifying thought switching points. Varying $\tau$ changes the granularity of thought segmentation and results in different trade-offs between reasoning accuracy and token efficiency. Nevertheless, Steady Thought remains consistently effective across all tested thresholds. As shown in Table~\ref{tab:threshlod-2}, on Qwen3-8B, varying $\tau$ from 1.0 to 1.6 yields MATH-500 accuracies between 92.6\% and 93.4\%, all higher than the 91.4\% baseline, while substantially reducing the number of output tokens. Similar trends are observed with DeepSeek-R1-Distill-Qwen-1.5B and across the other benchmarks. These results suggest that Steady Thought is not highly sensitive to the specific choice of $\tau$ within a reasonable range.

\begin{table*}[!t]

\caption{Results under varying thresholds.}
\label{tab:threshlod-2}
\renewcommand{\arraystretch}{0.9} 
\centering
\setlength\tabcolsep{3.5pt} 
\begin{adjustbox}{max width=\textwidth}
\begin{tabular}{@{}lcccccccc@{}}
\toprule
\multirow{2}{*}{Training model} & \multirow{2}{*}{Threshold}  & \multicolumn{2}{c}{MATH-500} & \multicolumn{2}{c}{AIME 2024} & \multicolumn{2}{c}{LiveCode} \\
&   & Acc(\%)$\uparrow$ & Tokens$\downarrow$ & Acc(\%)$\uparrow$ & Tokens$\downarrow$ & Acc(\%)$\uparrow$ & Tokens$\downarrow$ \\
\midrule
\multirow{1}{*}{DeepSeek-R1-Distill-Qwen-1.5B}
& -   & 82.0 & 4385 & 27.5 & 11273 & 30.3 & 9623 \\
\midrule
\multirow{3}{*}{+Steady Thought}
& 1.9   & 82.4 & 2987 & 27.5 & 7656 & 32.2 & 7843 \\

& 2.2   & 83.6 & 2691 & 30.8 & 7577 & 33.2 & 7335\\
& 2.5   & 82.6 & 2744 & 30.4 & 8889 & 32.8 & 7853\\
\midrule
\midrule
\multirow{1}{*}{Qwen3-8B}
& -  & 91.4 & 4724 & 62.1 & 10895 & 71.8 & 7112 \\
\midrule
\multirow{3}{*}{+Steady Thought}
& 1.0   & 92.6 & 3458 & 62.9 & 9623 & 83.0 & 6336\\
& 1.3   & 93.4 & 2139 & 64.6 & 7640 & 83.2 & 5631 \\
& 1.6   & 92.6 & 3548 & 62.9 & 9754 & 84.0 & 6297 \\
\bottomrule

\end{tabular}

\end{adjustbox}
\end{table*}

\section{Details of Data Construction}
\label{Construction}

\subsection{Computational Overhead of Data Construction}
\label{overhead}
We provide a detailed discussion on the computational overhead of Steady Thought, focusing on token consumption required for generating the preference data pairs during the trajectory synthesis for various model sizes as shown in Table \ref{tab:overhead_analysis}.

It should be noted that, for the 1.5B, 8B, 14B, and 32B models, trajectory synthesis was conducted using only a single 80GB GPU for inference acceleration. We consider this overhead to be acceptable and to yield a high return, particularly given its one-time nature and high efficiency.

\begin{table}[h]
\centering
\caption{Token consumption overview for trajectory synthesis stage}
\label{tab:overhead_analysis}
\begin{tabular}{l|c|c}
\toprule
Model & Total Tokens Consumed & Average Tokens Consumed \\
\midrule
DeepSeek-R1-Distill-Qwen-1.5B & 54304242 & 3356 \\
\midrule
Qwen3-8B & 104958704 & 6912 \\
\midrule
DeepSeek-R1-Distill-Qwen-14B & 45182779 & 3374 \\
\midrule
Qwen3-32B & 84345884 & 5441 \\
\bottomrule
\end{tabular}
\end{table}

\subsection{Data Utilization in Trajectory Synthesis}

To better characterize data utilization in the trajectory synthesis stage, we report the number of initial samples, segmented thoughts, and final converted preference pairs in Table~\ref{tab:data_utilization}. For each target model, we start from the same set of initial samples, segment the generated trajectories into thoughts, and then apply forced continuation to identify thoughts that can be successfully completed into correct answers.

\begin{table}[h]
\centering
\caption{Data utilization statistics in the Trajectory Synthesis stage.}
\label{tab:data_utilization}
\renewcommand{\arraystretch}{0.95} 
\setlength{\tabcolsep}{5pt}
\begin{tabular}{lccc}
\toprule
\textbf{Model} & \textbf{Initial Samples} & \textbf{Number of Thoughts} & \textbf{Converted Pairs} \\
\midrule
DeepSeek-R1-Distill-Qwen-1.5B & 1051 & 16177 & 2054 \\
Qwen3-8B & 1051 & 15185 & 1757 \\
DeepSeek-R1-Distill-Qwen-14B & 1051 & 13389 & 1842 \\
Qwen3-32B & 1051 & 15500 & 1877 \\
\bottomrule
\end{tabular}
\end{table}

We use successful forced continuation as a conservative criterion for determining whether the current thought should be preserved. Specifically, a thought is retained only when the model can continue from it and reach the correct final answer under the forced-continuation setting. Although this criterion may introduce false negatives by discarding some potentially viable thoughts that fail under constrained decoding, it helps ensure the cleanliness and reliability of the resulting preference data. To roughly estimate the false-negative rate, we randomly sample discarded thoughts and ask DeepSeek-V4-Pro to assess whether they remain on a potentially viable reasoning path given the question, reference answer, and reference solution. Approximately 8\% of the sampled discarded thoughts are judged to be potentially viable. Since these thoughts are discarded rather than included with incorrect labels, such false negatives primarily reduce data production efficiency rather than introduce noisy supervision into the retained preference data.

After applying this criterion and filtering out overly long instances, the conversion rate from segmented thoughts to final preference pairs is approximately 11.6\%--13.8\%. Moreover, since a single trajectory may contain multiple correct thoughts, one initial sample can yield multiple preference pairs. As a result, the number of converted pairs corresponds to approximately 167\%--195\% of the number of initial samples. These statistics show that the synthesis stage provides a substantial amount of usable supervision data in practice, and that potential false negatives do not substantially limit the availability of high-quality thought-level preference pairs.

\section{Example of Trajectory}
\label{example}
We demonstrate the stabilization effect brought by the ST by showing the Qwen3-8B model's response trajectory on the same problem before and after its application as presented in Table~\ref{segment case}. We also illustrate the segmentation results based on entropy. Observing the trajectory reveals that, prior to applying ST, the model frequently switched its thought direction on this problem, ultimately failing to complete the final answer within the pre-set maximum token limit. However, after applying ST, the number of thoughts was reduced from 11 to 4, the token count decreased from 14,336 to 7,413, and the answer was correct.

\begin{table}[t]
\caption{An example trajectory before and after ST.}
\centering
\small
\begin{tabular}{p{0.9\linewidth}}
\toprule
\textbf{Problem}: \\
The polynomial $x^3 - 3x^2 + 4x - 1$ is a factor of $x^9 + px^6 + qx^3 + r.$ Enter the ordered triple $(p,q,r).$ \\
\textbf{Reference Answer}: 
(6,31,-1)\\[0.5em]
\midrule
\textbf{\textsc{Before ST (14336 output tokens)}}:  \\
\begin{minipage}[t]{\linewidth}

Thought 1: Okay, so I need to find the ordered triple $(p, q, r)$ such that the polynomial
$x^3 - 3x^2 + 4x - 1$ is a factor of $x^9 + p x^6 + q x^3 + r.$ ...
Let me check if this makes sense. \\[0.5em]
Thought 2: Wait, let me verify this result by another method to be sure. Maybe by polynomial division or by considering that $x^9 \dots$ So perhaps my earlier approach is correct. \\[0.5em]
Thought 3: But let me check with the values I found. If $p = 6, \dots$ Let me check if $x^3 - 3x^2 + 4x - 1$ divides this. \\[0.5em]
Thought 4: Alternatively, let me try to perform the division \dots So there is a discrepancy in the coefficient of $p$. \\[0.5em]
Thought 5: Let me check the first approach again. \dots Wait, but this seems conflicting with the previous result. \\[0.5em]
Thought 6: Wait, actually, no. The original polynomial is $x^9 \dots$ Let me check my calculations again. \\[0.5em]
Thought 7: First, let me verify the first approach where I substituted $\alpha$ into the polynomial. \dots So the equations are correct. \\[0.5em]
Thought 8: But then why does the second approach give a different equation for the $x^2$ coefficient? Wait, maybe the second approach is not correct. \dots This seems correct. \\[0.5em]
Thought 9: The second approach, however, might have an error in the expansion or reduction. Let me recheck \dots there is a mistake in one of the approaches. \\[0.5em]
Thought 10: Wait, let me compute $\alpha^9 + p \alpha^6 + q \alpha^3 + r$ with $p=6$, $q=31$, $r=-1$ \dots Hence, this approach might not be straightforward. \\[0.5em]
Thought 11: Alternatively, if I substitute \(y=x^3\),
then \(P(x)=y^3+py^2+qy+r\).
But since \(x^3=3x^2-4x+1\),
then \(y=3x^2-4x+1\).
Therefore, \(P(x)=(3x^2-4x+1)^3+\cdots\)

\end{minipage} \\ 
\midrule
\textbf{\textsc{After ST (7413 output tokens)}}:  \\
\begin{minipage}[t]{\linewidth}

Thought 1: Okay, so I need to find the ordered triple $(p, q, r)$ such that the polynomial
$x^3 - 3x^2 + 4x - 1$ is a factor of $x^9 + p x^6 + q x^3 + r.$ \dots
This seems complicated, but maybe there's a smarter way. \\[0.5em]
Thought 2: Alternatively, since the given polynomial is a factor, then all roots \dots Let me check if indeed this is divisible by the cubic.\\[0.5em]
Thought 3: Alternatively, perhaps there is a better way. \dots Let me try this approach. \\[0.5em]
Thought 4: Let me assume that $Q(x) = x^6 + a x^3 + b. \dots$
\textbf{Final Answer}
The ordered triple is $\boxed{(6, 31, -1)}$. \\[0.5em]

\end{minipage} \\ 
\bottomrule

\end{tabular}
\label{segment case}
\end{table}

\section{Limitations}
\label{limitations}
Our evaluation covers multiple models and datasets, but it is still limited to a finite set of reasoning tasks. Future work could further examine the effectiveness of Steady Thought on broader model families and more diverse application scenarios. In addition, although our switching-point identification strategy is effective in practice, more adaptive detection mechanisms may further improve the efficiency of preference data construction.

\section{Broader Impacts}
\label{impacts}
This work improves the reasoning efficiency of large reasoning models by reducing redundant token generation while maintaining or improving accuracy. This may help lower inference cost, reduce energy consumption, and make advanced reasoning models more accessible in resource-constrained scenarios. However, since ST is built upon large language models, it may inherit their general risks.


\clearpage

\input{checklist.tex}

\end{document}

%% file: checklist.tex
\section*{NeurIPS Paper Checklist}

\begin{enumerate}

\item {\bf Claims}
    \item[] Question: Do the main claims made in the abstract and introduction accurately reflect the paper's contributions and scope?
    \item[] Answer: \answerYes{} 
    \item[] Justification: The abstract and introduction state the main contributions of ST, including thought-level preference optimization, token reduction, accuracy improvement, and generalization evaluation, which are supported by the experiments and analyses in the paper.
    \item[] Guidelines:
    \begin{itemize}
        \item The answer \answerNA{} means that the abstract and introduction do not include the claims made in the paper.
        \item The abstract and/or introduction should clearly state the claims made, including the contributions made in the paper and important assumptions and limitations. A \answerNo{} or \answerNA{} answer to this question will not be perceived well by the reviewers. 
        \item The claims made should match theoretical and experimental results, and reflect how much the results can be expected to generalize to other settings. 
        \item It is fine to include aspirational goals as motivation as long as it is clear that these goals are not attained by the paper. 
    \end{itemize}

\item {\bf Limitations}
    \item[] Question: Does the paper discuss the limitations of the work performed by the authors?
    \item[] Answer: \answerYes{} 
    \item[] Justification: The paper includes a limitations section discussing the finite scope of evaluated models and tasks and the efficiency of preference data construction (Appendix~\ref{limitations}).
    \item[] Guidelines:
    \begin{itemize}
        \item The answer \answerNA{} means that the paper has no limitation while the answer \answerNo{} means that the paper has limitations, but those are not discussed in the paper. 
        \item The authors are encouraged to create a separate ``Limitations'' section in their paper.
        \item The paper should point out any strong assumptions and how robust the results are to violations of these assumptions (e.g., independence assumptions, noiseless settings, model well-specification, asymptotic approximations only holding locally). The authors should reflect on how these assumptions might be violated in practice and what the implications would be.
        \item The authors should reflect on the scope of the claims made, e.g., if the approach was only tested on a few datasets or with a few runs. In general, empirical results often depend on implicit assumptions, which should be articulated.
        \item The authors should reflect on the factors that influence the performance of the approach. For example, a facial recognition algorithm may perform poorly when image resolution is low or images are taken in low lighting. Or a speech-to-text system might not be used reliably to provide closed captions for online lectures because it fails to handle technical jargon.
        \item The authors should discuss the computational efficiency of the proposed algorithms and how they scale with dataset size.
        \item If applicable, the authors should discuss possible limitations of their approach to address problems of privacy and fairness.
        \item While the authors might fear that complete honesty about limitations might be used by reviewers as grounds for rejection, a worse outcome might be that reviewers discover limitations that aren't acknowledged in the paper. The authors should use their best judgment and recognize that individual actions in favor of transparency play an important role in developing norms that preserve the integrity of the community. Reviewers will be specifically instructed to not penalize honesty concerning limitations.
    \end{itemize}

\item {\bf Theory assumptions and proofs}
    \item[] Question: For each theoretical result, does the paper provide the full set of assumptions and a complete (and correct) proof?
    \item[] Answer: \answerNA{} 
    \item[] Justification: The paper does not present formal theoretical theorems or proofs; the mathematical formulations are used to define the problem and optimization objective.
    \item[] Guidelines:
    \begin{itemize}
        \item The answer \answerNA{} means that the paper does not include theoretical results. 
        \item All the theorems, formulas, and proofs in the paper should be numbered and cross-referenced.
        \item All assumptions should be clearly stated or referenced in the statement of any theorems.
        \item The proofs can either appear in the main paper or the supplemental material, but if they appear in the supplemental material, the authors are encouraged to provide a short proof sketch to provide intuition. 
        \item Inversely, any informal proof provided in the core of the paper should be complemented by formal proofs provided in appendix or supplemental material.
        \item Theorems and Lemmas that the proof relies upon should be properly referenced. 
    \end{itemize}

    \item {\bf Experimental result reproducibility}
    \item[] Question: Does the paper fully disclose all the information needed to reproduce the main experimental results of the paper to the extent that it affects the main claims and/or conclusions of the paper (regardless of whether the code and data are provided or not)?
    \item[] Answer: \answerYes{} 
    \item[] Justification: The paper reports the training data source, evaluation datasets, baselines, hyperparameters, inference settings, and data construction procedure needed to reproduce the main experimental results.
    \item[] Guidelines:
    \begin{itemize}
        \item The answer \answerNA{} means that the paper does not include experiments.
        \item If the paper includes experiments, a \answerNo{} answer to this question will not be perceived well by the reviewers: Making the paper reproducible is important, regardless of whether the code and data are provided or not.
        \item If the contribution is a dataset and\slash or model, the authors should describe the steps taken to make their results reproducible or verifiable. 
        \item Depending on the contribution, reproducibility can be accomplished in various ways. For example, if the contribution is a novel architecture, describing the architecture fully might suffice, or if the contribution is a specific model and empirical evaluation, it may be necessary to either make it possible for others to replicate the model with the same dataset, or provide access to the model. In general. releasing code and data is often one good way to accomplish this, but reproducibility can also be provided via detailed instructions for how to replicate the results, access to a hosted model (e.g., in the case of a large language model), releasing of a model checkpoint, or other means that are appropriate to the research performed.
        \item While NeurIPS does not require releasing code, the conference does require all submissions to provide some reasonable avenue for reproducibility, which may depend on the nature of the contribution. For example
        \begin{enumerate}
            \item If the contribution is primarily a new algorithm, the paper should make it clear how to reproduce that algorithm.
            \item If the contribution is primarily a new model architecture, the paper should describe the architecture clearly and fully.
            \item If the contribution is a new model (e.g., a large language model), then there should either be a way to access this model for reproducing the results or a way to reproduce the model (e.g., with an open-source dataset or instructions for how to construct the dataset).
            \item We recognize that reproducibility may be tricky in some cases, in which case authors are welcome to describe the particular way they provide for reproducibility. In the case of closed-source models, it may be that access to the model is limited in some way (e.g., to registered users), but it should be possible for other researchers to have some path to reproducing or verifying the results.
        \end{enumerate}
    \end{itemize}

\item {\bf Open access to data and code}
    \item[] Question: Does the paper provide open access to the data and code, with sufficient instructions to faithfully reproduce the main experimental results, as described in supplemental material?
    \item[] Answer: \answerYes{} 
    \item[] Justification: We use the open-source datasets and benchmarks, and the code used in this paper has been made publicly available.
    \item[] Guidelines:
    \begin{itemize}
        \item The answer \answerNA{} means that paper does not include experiments requiring code.
        \item Please see the NeurIPS code and data submission guidelines (\url{https://neurips.cc/public/guides/CodeSubmissionPolicy}) for more details.
        \item While we encourage the release of code and data, we understand that this might not be possible, so \answerNo{} is an acceptable answer. Papers cannot be rejected simply for not including code, unless this is central to the contribution (e.g., for a new open-source benchmark).
        \item The instructions should contain the exact command and environment needed to run to reproduce the results. See the NeurIPS code and data submission guidelines (\url{https://neurips.cc/public/guides/CodeSubmissionPolicy}) for more details.
        \item The authors should provide instructions on data access and preparation, including how to access the raw data, preprocessed data, intermediate data, and generated data, etc.
        \item The authors should provide scripts to reproduce all experimental results for the new proposed method and baselines. If only a subset of experiments are reproducible, they should state which ones are omitted from the script and why.
        \item At submission time, to preserve anonymity, the authors should release anonymized versions (if applicable).
        \item Providing as much information as possible in supplemental material (appended to the paper) is recommended, but including URLs to data and code is permitted.
    \end{itemize}

\item {\bf Experimental setting/details}
    \item[] Question: Does the paper specify all the training and test details (e.g., data splits, hyperparameters, how they were chosen, type of optimizer) necessary to understand the results?
    \item[] Answer: \answerYes{} 
    \item[] Justification: The paper specifies the training datasets, evaluation benchmarks, decoding parameters, optimization settings, entropy thresholds, and model-specific training hyperparameters in the main text and appendix.
    \item[] Guidelines:
    \begin{itemize}
        \item The answer \answerNA{} means that the paper does not include experiments.
        \item The experimental setting should be presented in the core of the paper to a level of detail that is necessary to appreciate the results and make sense of them.
        \item The full details can be provided either with the code, in appendix, or as supplemental material.
    \end{itemize}

\item {\bf Experiment statistical significance}
    \item[] Question: Does the paper report error bars suitably and correctly defined or other appropriate information about the statistical significance of the experiments?
    \item[] Answer: \answerYes{} 
    \item[] Justification: We report averaged results over multiple evaluation runs for benchmarks with higher variability, including eight runs for AIME 2024 and two runs for LiveCode. For MATH500 and GSM8K, we use a single run because these datasets contain larger test sets and exhibit relatively stable model outputs, as specified in the experimental protocol.
    \item[] Guidelines:
    \begin{itemize}
        \item The answer \answerNA{} means that the paper does not include experiments.
        \item The authors should answer \answerYes{} if the results are accompanied by error bars, confidence intervals, or statistical significance tests, at least for the experiments that support the main claims of the paper.
        \item The factors of variability that the error bars are capturing should be clearly stated (for example, train/test split, initialization, random drawing of some parameter, or overall run with given experimental conditions).
        \item The method for calculating the error bars should be explained (closed form formula, call to a library function, bootstrap, etc.)
        \item The assumptions made should be given (e.g., Normally distributed errors).
        \item It should be clear whether the error bar is the standard deviation or the standard error of the mean.
        \item It is OK to report 1-sigma error bars, but one should state it. The authors should preferably report a 2-sigma error bar than state that they have a 96\% CI, if the hypothesis of Normality of errors is not verified.
        \item For asymmetric distributions, the authors should be careful not to show in tables or figures symmetric error bars that would yield results that are out of range (e.g., negative error rates).
        \item If error bars are reported in tables or plots, the authors should explain in the text how they were calculated and reference the corresponding figures or tables in the text.
    \end{itemize}

\item {\bf Experiments compute resources}
    \item[] Question: For each experiment, does the paper provide sufficient information on the computer resources (type of compute workers, memory, time of execution) needed to reproduce the experiments?
    \item[] Answer: \answerYes{} 
    \item[] Justification: The paper reports the main training and inference settings and provides additional computational overhead statistics for trajectory synthesis and preference data construction in the appendix.
    \item[] Guidelines:
    \begin{itemize}
        \item The answer \answerNA{} means that the paper does not include experiments.
        \item The paper should indicate the type of compute workers CPU or GPU, internal cluster, or cloud provider, including relevant memory and storage.
        \item The paper should provide the amount of compute required for each of the individual experimental runs as well as estimate the total compute. 
        \item The paper should disclose whether the full research project required more compute than the experiments reported in the paper (e.g., preliminary or failed experiments that didn't make it into the paper). 
    \end{itemize}
    
\item {\bf Code of ethics}
    \item[] Question: Does the research conducted in the paper conform, in every respect, with the NeurIPS Code of Ethics \url{https://neurips.cc/public/EthicsGuidelines}?
    \item[] Answer: \answerYes{} 
    \item[] Justification: The work uses publicly available models and benchmarks for reasoning research and does not involve private data, human-subject experiments, or procedures that deviate from the NeurIPS Code of Ethics.
    \item[] Guidelines:
    \begin{itemize}
        \item The answer \answerNA{} means that the authors have not reviewed the NeurIPS Code of Ethics.
        \item If the authors answer \answerNo, they should explain the special circumstances that require a deviation from the Code of Ethics.
        \item The authors should make sure to preserve anonymity (e.g., if there is a special consideration due to laws or regulations in their jurisdiction).
    \end{itemize}

\item {\bf Broader impacts}
    \item[] Question: Does the paper discuss both potential positive societal impacts and negative societal impacts of the work performed?
    \item[] Answer: \answerYes{} 
    \item[] Justification: The paper discusses that improving reasoning efficiency can reduce computational cost, while also acknowledging that more efficient reasoning models may inherit the general risks associated with powerful language models (Appendix~\ref{impacts}).
    \item[] Guidelines:
    \begin{itemize}
        \item The answer \answerNA{} means that there is no societal impact of the work performed.
        \item If the authors answer \answerNA{} or \answerNo, they should explain why their work has no societal impact or why the paper does not address societal impact.
        \item The conference expects that many papers will be foundational research and not tied to particular applications, let alone deployments. However, if there is a direct path to any negative applications, the authors should point it out. For example, it is legitimate to point out that an improvement in the quality of generative models could be used to generate Deepfakes for disinformation. On the other hand, it is not needed to point out that a generic algorithm for optimizing neural networks could enable people to train models that generate Deepfakes faster.
        \item The authors should consider possible harms that could arise when the technology is being used as intended and functioning correctly, harms that could arise when the technology is being used as intended but gives incorrect results, and harms following from (intentional or unintentional) misuse of the technology.
        \item If there are negative societal impacts, the authors could also discuss possible mitigation strategies (e.g., gated release of models, providing defenses in addition to attacks, mechanisms for monitoring misuse, mechanisms to monitor how a system learns from feedback over time, improving the efficiency and accessibility of ML).
    \end{itemize}
    
\item {\bf Safeguards}
    \item[] Question: Does the paper describe safeguards that have been put in place for responsible release of data or models that have a high risk for misuse (e.g., pre-trained language models, image generators, or scraped datasets)?
    \item[] Answer: \answerNA{} 
    \item[] Justification: The paper does not release a new high-risk pretrained model, scraped dataset, or generative system requiring additional safeguards beyond those associated with the original public models and datasets.
    \item[] Guidelines:
    \begin{itemize}
        \item The answer \answerNA{} means that the paper poses no such risks.
        \item Released models that have a high risk for misuse or dual-use should be released with necessary safeguards to allow for controlled use of the model, for example by requiring that users adhere to usage guidelines or restrictions to access the model or implementing safety filters. 
        \item Datasets that have been scraped from the Internet could pose safety risks. The authors should describe how they avoided releasing unsafe images.
        \item We recognize that providing effective safeguards is challenging, and many papers do not require this, but we encourage authors to take this into account and make a best faith effort.
    \end{itemize}

\item {\bf Licenses for existing assets}
    \item[] Question: Are the creators or original owners of assets (e.g., code, data, models), used in the paper, properly credited and are the license and terms of use explicitly mentioned and properly respected?
    \item[] Answer: \answerYes{} 
    \item[] Justification: The paper cites the original sources of the models, datasets, and baselines used in the experiments, and these assets are used for research evaluation consistent with their intended academic use.
    \item[] Guidelines:
    \begin{itemize}
        \item The answer \answerNA{} means that the paper does not use existing assets.
        \item The authors should cite the original paper that produced the code package or dataset.
        \item The authors should state which version of the asset is used and, if possible, include a URL.
        \item The name of the license (e.g., CC-BY 4.0) should be included for each asset.
        \item For scraped data from a particular source (e.g., website), the copyright and terms of service of that source should be provided.
        \item If assets are released, the license, copyright information, and terms of use in the package should be provided. For popular datasets, \url{paperswithcode.com/datasets} has curated licenses for some datasets. Their licensing guide can help determine the license of a dataset.
        \item For existing datasets that are re-packaged, both the original license and the license of the derived asset (if it has changed) should be provided.
        \item If this information is not available online, the authors are encouraged to reach out to the asset's creators.
    \end{itemize}

\item {\bf New assets}
    \item[] Question: Are new assets introduced in the paper well documented and is the documentation provided alongside the assets?
    \item[] Answer: \answerNA{} 
    \item[] Justification: This paper does not release new assets at submission time; the generated preference data and code will be documented and released upon acceptance.
    \item[] Guidelines:
    \begin{itemize}
        \item The answer \answerNA{} means that the paper does not release new assets.
        \item Researchers should communicate the details of the dataset\slash code\slash model as part of their submissions via structured templates. This includes details about training, license, limitations, etc. 
        \item The paper should discuss whether and how consent was obtained from people whose asset is used.
        \item At submission time, remember to anonymize your assets (if applicable). You can either create an anonymized URL or include an anonymized zip file.
    \end{itemize}

\item {\bf Crowdsourcing and research with human subjects}
    \item[] Question: For crowdsourcing experiments and research with human subjects, does the paper include the full text of instructions given to participants and screenshots, if applicable, as well as details about compensation (if any)? 
    \item[] Answer: \answerNA{} 
    \item[] Justification: The paper does not involve crowdsourcing experiments or research with human subjects.
    \item[] Guidelines:
    \begin{itemize}
        \item The answer \answerNA{} means that the paper does not involve crowdsourcing nor research with human subjects.
        \item Including this information in the supplemental material is fine, but if the main contribution of the paper involves human subjects, then as much detail as possible should be included in the main paper. 
        \item According to the NeurIPS Code of Ethics, workers involved in data collection, curation, or other labor should be paid at least the minimum wage in the country of the data collector. 
    \end{itemize}

\item {\bf Institutional review board (IRB) approvals or equivalent for research with human subjects}
    \item[] Question: Does the paper describe potential risks incurred by study participants, whether such risks were disclosed to the subjects, and whether Institutional Review Board (IRB) approvals (or an equivalent approval/review based on the requirements of your country or institution) were obtained?
    \item[] Answer: \answerNA{} 
    \item[] Justification: The paper does not involve human subjects, so IRB approval or equivalent review is not applicable.
    \item[] Guidelines:
    \begin{itemize}
        \item The answer \answerNA{} means that the paper does not involve crowdsourcing nor research with human subjects.
        \item Depending on the country in which research is conducted, IRB approval (or equivalent) may be required for any human subjects research. If you obtained IRB approval, you should clearly state this in the paper. 
        \item We recognize that the procedures for this may vary significantly between institutions and locations, and we expect authors to adhere to the NeurIPS Code of Ethics and the guidelines for their institution. 
        \item For initial submissions, do not include any information that would break anonymity (if applicable), such as the institution conducting the review.
    \end{itemize}

\item {\bf Declaration of LLM usage}
    \item[] Question: Does the paper describe the usage of LLMs if it is an important, original, or non-standard component of the core methods in this research? Note that if the LLM is used only for writing, editing, or formatting purposes and does \emph{not} impact the core methodology, scientific rigor, or originality of the research, declaration is not required.
    \item[] Answer: \answerYes{} 
    \item[] Justification: The core method directly studies and trains large reasoning models, and the paper describes how these models are used for trajectory generation, forced continuation, and evaluation.
    \item[] Guidelines:
    \begin{itemize}
        \item The answer \answerNA{} means that the core method development in this research does not involve LLMs as any important, original, or non-standard components.
        \item Please refer to our LLM policy in the NeurIPS handbook for what should or should not be described.
    \end{itemize}

\end{enumerate}

%% file: reference.bib
@inproceedings{
wang2025logictree,
title={LogicTree: Improving Complex Reasoning of {LLM}s via Instantiated Multi-step Synthetic Logical Data},
author={Zehao Wang and Lin Yang and Jie Wang and Kehan Wang and Hanzhu Chen and Bin Wang and Jianye HAO and Defu Lian and Bin Li and Enhong Chen},
booktitle={The Thirty-ninth Annual Conference on Neural Information Processing Systems},
year={2025},
url={https://openreview.net/forum?id=z4AMrCOetn}
}

@inproceedings{liang2026boosting,
  title={Boosting Multi-Domain Reasoning of LLMs via Curvature-Guided Policy Optimization},
  author={Liang, Xize and Yang, Lin and Wang, Jie and Liu, Rui and Lu, Yang and Zeng, Jinliang and Chen, Hanzhu and Li, Dong and Hao, Jianye},
  booktitle={The Fourteenth International Conference on Learning Representations},
  year={2026}
}

@inproceedings{liang2025boosting,
  title={Boosting Multi-Domain Fine-Tuning of Large Language Models through Evolving Interactions between Samples},
  author={Liang, Xize and Yang, Lin and Wang, Jie and Lu, Yiyang and Wu, Runyu and Chen, Hanzhu and Hao, Jianye},
  booktitle={International Conference on Machine Learning},
  pages={37427--37441},
  year={2025},
  organization={PMLR}
}

@inproceedings{liang2025ropo,
  title={ROPO: Robust Preference Optimization for Large Language Models},
  author={Liang, Xize and Chen, Chao and Qiu, Shuang and Wang, Jie and Wu, Yue and Fu, Zhihang and Chen, Hanzhu and Wu, Feng and Ye, Jieping},
  booktitle={International Conference on Machine Learning},
  pages={37131--37161},
  year={2025},
  organization={PMLR}
}

@inproceedings{chen2026latent,
  title={Latent-Guided Reasoning: Empowering Small LLMs with Large-Model Thinking},
  author={Chen, Hanzhu and Yang, Lin and Wang, Jie and Yan, Junhao and Wang, Zhe and Liang, Xize and Hao, Jianye},
  booktitle={The Fourteenth International Conference on Learning Representations},
  year={2026}
}

@inproceedings{ma2026styletailor,
  title={Styletailor: Towards personalized fashion styling via hierarchical negative feedback},
  author={Ma, Hongbo and Shen, Fei and Xu, Hongbin and Wang, Xiaoce and Xu, Gang and Zheng, Jinkai and Qu, Liangqiong and Li, Ming},
  booktitle={Proceedings of the AAAI Conference on Artificial Intelligence},
  volume={40},
  number={35},
  pages={29591--29599},
  year={2026}
}

@misc{cobbe2021trainingverifierssolvemath,
      title={Training Verifiers to Solve Math Word Problems}, 
      author={Karl Cobbe and Vineet Kosaraju and Mohammad Bavarian and Mark Chen and Heewoo Jun and Lukasz Kaiser and Matthias Plappert and Jerry Tworek and Jacob Hilton and Reiichiro Nakano and Christopher Hesse and John Schulman},
      year={2021},
      eprint={2110.14168},
      archivePrefix={arXiv},
      primaryClass={cs.LG},
      url={https://arxiv.org/abs/2110.14168}, 
}

@misc{nayab2025concisethoughtsimpactoutput,
      title={Concise Thoughts: Impact of Output Length on LLM Reasoning and Cost}, 
      author={Sania Nayab and Giulio Rossolini and Marco Simoni and Andrea Saracino and Giorgio Buttazzo and Nicolamaria Manes and Fabrizio Giacomelli},
      year={2025},
      eprint={2407.19825},
      archivePrefix={arXiv},
      primaryClass={cs.CL},
      url={https://arxiv.org/abs/2407.19825}, 
}

@misc{xu2025chaindraftthinkingfaster,
      title={Chain of Draft: Thinking Faster by Writing Less}, 
      author={Silei Xu and Wenhao Xie and Lingxiao Zhao and Pengcheng He},
      year={2025},
      eprint={2502.18600},
      archivePrefix={arXiv},
      primaryClass={cs.CL},
      url={https://arxiv.org/abs/2502.18600}, 
}

@misc{sui2025stopoverthinkingsurveyefficient,
      title={Stop Overthinking: A Survey on Efficient Reasoning for Large Language Models}, 
      author={Yang Sui and Yu-Neng Chuang and Guanchu Wang and Jiamu Zhang and Tianyi Zhang and Jiayi Yuan and Hongyi Liu and Andrew Wen and Shaochen Zhong and Hanjie Chen and Xia Hu},
      year={2025},
      eprint={2503.16419},
      archivePrefix={arXiv},
      primaryClass={cs.CL},
      url={https://arxiv.org/abs/2503.16419}, 
}

@misc{qu2025surveyefficientreasoninglarge,
      title={A Survey of Efficient Reasoning for Large Reasoning Models: Language, Multimodality, and Beyond}, 
      author={Xiaoye Qu and Yafu Li and Zhaochen Su and Weigao Sun and Jianhao Yan and Dongrui Liu and Ganqu Cui and Daizong Liu and Shuxian Liang and Junxian He and Peng Li and Wei Wei and Jing Shao and Chaochao Lu and Yue Zhang and Xian-Sheng Hua and Bowen Zhou and Yu Cheng},
      year={2025},
      eprint={2503.21614},
      archivePrefix={arXiv},
      primaryClass={cs.CL},
      url={https://arxiv.org/abs/2503.21614}, 
}

@misc{qiu2024treebonenhancinginferencetimealignment,
      title={TreeBoN: Enhancing Inference-Time Alignment with Speculative Tree-Search and Best-of-N Sampling}, 
      author={Jiahao Qiu and Yifu Lu and Yifan Zeng and Jiacheng Guo and Jiayi Geng and Huazheng Wang and Kaixuan Huang and Yue Wu and Mengdi Wang},
      year={2024},
      eprint={2410.16033},
      archivePrefix={arXiv},
      primaryClass={cs.CL},
      url={https://arxiv.org/abs/2410.16033}, 
}

@misc{zhang2025lightthinkerthinkingstepbystepcompression,
      title={LightThinker: Thinking Step-by-Step Compression}, 
      author={Jintian Zhang and Yuqi Zhu and Mengshu Sun and Yujie Luo and Shuofei Qiao and Lun Du and Da Zheng and Huajun Chen and Ningyu Zhang},
      year={2025},
      eprint={2502.15589},
      archivePrefix={arXiv},
      primaryClass={cs.CL},
      url={https://arxiv.org/abs/2502.15589}, 
}

@misc{kang2024c3otgeneratingshorterchainofthought,
      title={C3oT: Generating Shorter Chain-of-Thought without Compromising Effectiveness}, 
      author={Yu Kang and Xianghui Sun and Liangyu Chen and Wei Zou},
      year={2024},
      eprint={2412.11664},
      archivePrefix={arXiv},
      primaryClass={cs.CL},
      url={https://arxiv.org/abs/2412.11664}, 
}

@misc{aggarwal2025l1controllinglongreasoning,
      title={L1: Controlling How Long A Reasoning Model Thinks With Reinforcement Learning}, 
      author={Pranjal Aggarwal and Sean Welleck},
      year={2025},
      eprint={2503.04697},
      archivePrefix={arXiv},
      primaryClass={cs.CL},
      url={https://arxiv.org/abs/2503.04697}, 
}

@misc{zhang2025adaptthinkreasoningmodelslearn,
      title={AdaptThink: Reasoning Models Can Learn When to Think}, 
      author={Jiajie Zhang and Nianyi Lin and Lei Hou and Ling Feng and Juanzi Li},
      year={2025},
      eprint={2505.13417},
      archivePrefix={arXiv},
      primaryClass={cs.CL},
      url={https://arxiv.org/abs/2505.13417}, 
}

@misc{lou2025adacotparetooptimaladaptivechainofthought,
      title={AdaCoT: Pareto-Optimal Adaptive Chain-of-Thought Triggering via Reinforcement Learning}, 
      author={Chenwei Lou and Zewei Sun and Xinnian Liang and Meng Qu and Wei Shen and Wenqi Wang and Yuntao Li and Qingping Yang and Shuangzhi Wu},
      year={2025},
      eprint={2505.11896},
      archivePrefix={arXiv},
      primaryClass={cs.LG},
      url={https://arxiv.org/abs/2505.11896}, 
}

@misc{chen2025sealsteerablereasoningcalibration,
      title={SEAL: Steerable Reasoning Calibration of Large Language Models for Free}, 
      author={Runjin Chen and Zhenyu Zhang and Junyuan Hong and Souvik Kundu and Zhangyang Wang},
      year={2025},
      eprint={2504.07986},
      archivePrefix={arXiv},
      primaryClass={cs.CL},
      url={https://arxiv.org/abs/2504.07986}, 
}

@misc{wang2025thoughtsplaceunderthinkingo1like,
      title={Thoughts Are All Over the Place: On the Underthinking of o1-Like LLMs}, 
      author={Yue Wang and Qiuzhi Liu and Jiahao Xu and Tian Liang and Xingyu Chen and Zhiwei He and Linfeng Song and Dian Yu and Juntao Li and Zhuosheng Zhang and Rui Wang and Zhaopeng Tu and Haitao Mi and Dong Yu},
      year={2025},
      eprint={2501.18585},
      archivePrefix={arXiv},
      primaryClass={cs.CL},
      url={https://arxiv.org/abs/2501.18585}, 
}

@misc{wang2025waitdontneedwait,
      title={Wait, We Don't Need to "Wait"! Removing Thinking Tokens Improves Reasoning Efficiency}, 
      author={Chenlong Wang and Yuanning Feng and Dongping Chen and Zhaoyang Chu and Ranjay Krishna and Tianyi Zhou},
      year={2025},
      eprint={2506.08343},
      archivePrefix={arXiv},
      primaryClass={cs.CL},
      url={https://arxiv.org/abs/2506.08343}, 
}

@misc{ding2025thinkingtokenshelptrap,
      title={Do Thinking Tokens Help or Trap? Towards More Efficient Large Reasoning Model}, 
      author={Bowen Ding and Yuhan Chen and Futing Wang and Lingfeng Ming and Tao Lin},
      year={2025},
      eprint={2506.23840},
      archivePrefix={arXiv},
      primaryClass={cs.CL},
      url={https://arxiv.org/abs/2506.23840}, 
}

@misc{deepseekai2025deepseekr1incentivizingreasoningcapability,
      title={DeepSeek-R1: Incentivizing Reasoning Capability in LLMs via Reinforcement Learning}, 
      author={DeepSeek-AI and Daya Guo and Dejian Yang and Haowei Zhang and Junxiao Song and Ruoyu Zhang and Runxin Xu and Qihao Zhu and Shirong Ma and Peiyi Wang and Xiao Bi and Xiaokang Zhang and Xingkai Yu and Yu Wu and Z. F. Wu and Zhibin Gou and Zhihong Shao and Zhuoshu Li and Ziyi Gao and Aixin Liu and Bing Xue and Bingxuan Wang and Bochao Wu and Bei Feng and Chengda Lu and Chenggang Zhao and Chengqi Deng and Chenyu Zhang and Chong Ruan and Damai Dai and Deli Chen and Dongjie Ji and Erhang Li and Fangyun Lin and Fucong Dai and Fuli Luo and Guangbo Hao and Guanting Chen and Guowei Li and H. Zhang and Han Bao and Hanwei Xu and Haocheng Wang and Honghui Ding and Huajian Xin and Huazuo Gao and Hui Qu and Hui Li and Jianzhong Guo and Jiashi Li and Jiawei Wang and Jingchang Chen and Jingyang Yuan and Junjie Qiu and Junlong Li and J. L. Cai and Jiaqi Ni and Jian Liang and Jin Chen and Kai Dong and Kai Hu and Kaige Gao and Kang Guan and Kexin Huang and Kuai Yu and Lean Wang and Lecong Zhang and Liang Zhao and Litong Wang and Liyue Zhang and Lei Xu and Leyi Xia and Mingchuan Zhang and Minghua Zhang and Minghui Tang and Meng Li and Miaojun Wang and Mingming Li and Ning Tian and Panpan Huang and Peng Zhang and Qiancheng Wang and Qinyu Chen and Qiushi Du and Ruiqi Ge and Ruisong Zhang and Ruizhe Pan and Runji Wang and R. J. Chen and R. L. Jin and Ruyi Chen and Shanghao Lu and Shangyan Zhou and Shanhuang Chen and Shengfeng Ye and Shiyu Wang and Shuiping Yu and Shunfeng Zhou and Shuting Pan and S. S. Li and Shuang Zhou and Shaoqing Wu and Shengfeng Ye and Tao Yun and Tian Pei and Tianyu Sun and T. Wang and Wangding Zeng and Wanjia Zhao and Wen Liu and Wenfeng Liang and Wenjun Gao and Wenqin Yu and Wentao Zhang and W. L. Xiao and Wei An and Xiaodong Liu and Xiaohan Wang and Xiaokang Chen and Xiaotao Nie and Xin Cheng and Xin Liu and Xin Xie and Xingchao Liu and Xinyu Yang and Xinyuan Li and Xuecheng Su and Xuheng Lin and X. Q. Li and Xiangyue Jin and Xiaojin Shen and Xiaosha Chen and Xiaowen Sun and Xiaoxiang Wang and Xinnan Song and Xinyi Zhou and Xianzu Wang and Xinxia Shan and Y. K. Li and Y. Q. Wang and Y. X. Wei and Yang Zhang and Yanhong Xu and Yao Li and Yao Zhao and Yaofeng Sun and Yaohui Wang and Yi Yu and Yichao Zhang and Yifan Shi and Yiliang Xiong and Ying He and Yishi Piao and Yisong Wang and Yixuan Tan and Yiyang Ma and Yiyuan Liu and Yongqiang Guo and Yuan Ou and Yuduan Wang and Yue Gong and Yuheng Zou and Yujia He and Yunfan Xiong and Yuxiang Luo and Yuxiang You and Yuxuan Liu and Yuyang Zhou and Y. X. Zhu and Yanhong Xu and Yanping Huang and Yaohui Li and Yi Zheng and Yuchen Zhu and Yunxian Ma and Ying Tang and Yukun Zha and Yuting Yan and Z. Z. Ren and Zehui Ren and Zhangli Sha and Zhe Fu and Zhean Xu and Zhenda Xie and Zhengyan Zhang and Zhewen Hao and Zhicheng Ma and Zhigang Yan and Zhiyu Wu and Zihui Gu and Zijia Zhu and Zijun Liu and Zilin Li and Ziwei Xie and Ziyang Song and Zizheng Pan and Zhen Huang and Zhipeng Xu and Zhongyu Zhang and Zhen Zhang},
      year={2025},
      eprint={2501.12948},
      archivePrefix={arXiv},
      primaryClass={cs.CL},
      url={https://arxiv.org/abs/2501.12948}, 
}

@misc{o1,
    title = {Learning to Reason with LLMs},
    url = {https://openai.com/index/learning-to-reason-with-llms/},
    author = {OpenAI},
    month = {September},
    year = {2024},
}

@misc{muennighoff2025s1simpletesttimescaling,
      title={s1: Simple test-time scaling}, 
      author={Niklas Muennighoff and Zitong Yang and Weijia Shi and Xiang Lisa Li and Li Fei-Fei and Hannaneh Hajishirzi and Luke Zettlemoyer and Percy Liang and Emmanuel Candès and Tatsunori Hashimoto},
      year={2025},
      eprint={2501.19393},
      archivePrefix={arXiv},
      primaryClass={cs.CL},
      url={https://arxiv.org/abs/2501.19393}, 
}

@misc{zeng2025simplerlzooinvestigatingtamingzero,
      title={SimpleRL-Zoo: Investigating and Taming Zero Reinforcement Learning for Open Base Models in the Wild}, 
      author={Weihao Zeng and Yuzhen Huang and Qian Liu and Wei Liu and Keqing He and Zejun Ma and Junxian He},
      year={2025},
      eprint={2503.18892},
      archivePrefix={arXiv},
      primaryClass={cs.LG},
      url={https://arxiv.org/abs/2503.18892}, 
}

@misc{chen2025empiricalstudyelicitingimproving,
      title={An Empirical Study on Eliciting and Improving R1-like Reasoning Models}, 
      author={Zhipeng Chen and Yingqian Min and Beichen Zhang and Jie Chen and Jinhao Jiang and Daixuan Cheng and Wayne Xin Zhao and Zheng Liu and Xu Miao and Yang Lu and Lei Fang and Zhongyuan Wang and Ji-Rong Wen},
      year={2025},
      eprint={2503.04548},
      archivePrefix={arXiv},
      primaryClass={cs.CL},
      url={https://arxiv.org/abs/2503.04548}, 
}

@misc{yang2025understandingahamomentsexternal,
      title={Understanding Aha Moments: from External Observations to Internal Mechanisms}, 
      author={Shu Yang and Junchao Wu and Xin Chen and Yunze Xiao and Xinyi Yang and Derek F. Wong and Di Wang},
      year={2025},
      eprint={2504.02956},
      archivePrefix={arXiv},
      primaryClass={cs.CL},
      url={https://arxiv.org/abs/2504.02956}, 
}

@misc{meng2024simposimplepreferenceoptimization,
      title={SimPO: Simple Preference Optimization with a Reference-Free Reward}, 
      author={Yu Meng and Mengzhou Xia and Danqi Chen},
      year={2024},
      eprint={2405.14734},
      archivePrefix={arXiv},
      primaryClass={cs.CL},
      url={https://arxiv.org/abs/2405.14734}, 
}

@misc{gao2024omnimathuniversalolympiadlevel,
      title={Omni-MATH: A Universal Olympiad Level Mathematic Benchmark For Large Language Models}, 
      author={Bofei Gao and Feifan Song and Zhe Yang and Zefan Cai and Yibo Miao and Qingxiu Dong and Lei Li and Chenghao Ma and Liang Chen and Runxin Xu and Zhengyang Tang and Benyou Wang and Daoguang Zan and Shanghaoran Quan and Ge Zhang and Lei Sha and Yichang Zhang and Xuancheng Ren and Tianyu Liu and Baobao Chang},
      year={2024},
      eprint={2410.07985},
      archivePrefix={arXiv},
      primaryClass={cs.CL},
      url={https://arxiv.org/abs/2410.07985}, 
}

@misc{qwen3-2024,
  author       = {{Qwen Team}},
  title        = {{Qwen3}: {A} Series of Large Language Models},
  howpublished = {\url{https://qwenlm.github.io/blog/qwen3/}},
  year         = {2025}
}

@misc{hendrycks2021measuringmathematicalproblemsolving,
      title={Measuring Mathematical Problem Solving With the MATH Dataset}, 
      author={Dan Hendrycks and Collin Burns and Saurav Kadavath and Akul Arora and Steven Basart and Eric Tang and Dawn Song and Jacob Steinhardt},
      year={2021},
      eprint={2103.03874},
      archivePrefix={arXiv},
      primaryClass={cs.LG},
      url={https://arxiv.org/abs/2103.03874}, 
}

@misc{jain2024livecodebenchholisticcontaminationfree,
      title={LiveCodeBench: Holistic and Contamination Free Evaluation of Large Language Models for Code}, 
      author={Naman Jain and King Han and Alex Gu and Wen-Ding Li and Fanjia Yan and Tianjun Zhang and Sida Wang and Armando Solar-Lezama and Koushik Sen and Ion Stoica},
      year={2024},
      eprint={2403.07974},
      archivePrefix={arXiv},
      primaryClass={cs.SE},
      url={https://arxiv.org/abs/2403.07974}, 
}

@misc{maa-aime-2024,
  author       = {{MAA Committees}},
  title        = {{AIME} Problems and Solutions},
  howpublished = {\url{https://artofproblemsolving.com/wiki/index.php/AIME_Problems_and_Solutions}},
  year         = {2024},
  note         = {Accessed: 2025-09-23}
}

@misc{ma2025reasoningmodelseffectivethinking,
      title={Reasoning Models Can Be Effective Without Thinking}, 
      author={Wenjie Ma and Jingxuan He and Charlie Snell and Tyler Griggs and Sewon Min and Matei Zaharia},
      year={2025},
      eprint={2504.09858},
      archivePrefix={arXiv},
      primaryClass={cs.AI},
      url={https://arxiv.org/abs/2504.09858}, 
}

@article{rafailov2024direct,
  title={Direct preference optimization: Your language model is secretly a reward model},
  author={Rafailov, Rafael and Sharma, Archit and Mitchell, Eric and Manning, Christopher D and Ermon, Stefano and Finn, Chelsea},
  journal={Advances in Neural Information Processing Systems},
  volume={36},
  year={2024}
}

@inproceedings{lightmanlet,
  title={Let's Verify Step by Step},
  author={Lightman, Hunter and Kosaraju, Vineet and Burda, Yuri and Edwards, Harrison and Baker, Bowen and Lee, Teddy and Leike, Jan and Schulman, John and Sutskever, Ilya and Cobbe, Karl},
  booktitle={The Twelfth International Conference on Learning Representations},
  year={2024}
}

@article{luo2023wizardmath,
  title={Wizardmath: Empowering mathematical reasoning for large language models via reinforced evol-instruct},
  author={Luo, Haipeng and Sun, Qingfeng and Xu, Can and Zhao, Pu and Lou, Jianguang and Tao, Chongyang and Geng, Xiubo and Lin, Qingwei and Chen, Shifeng and Zhang, Dongmei},
  journal={arXiv:2308.09583},
  year={2023}
}

@misc{lu2024stepcontrolled,
    title={Step-Controlled DPO: Leveraging Stepwise Error for Enhanced Mathematical Reasoning},
    author={Zimu Lu and Aojun Zhou and Ke Wang and Houxing Ren and Weikang Shi and Junting Pan and Mingjie Zhan and Hongsheng Li},
    year={2024},
    eprint={2407.00782},
    archivePrefix={arXiv},
    primaryClass={cs.CL}
}

@misc{lai2024stepdpo,
    title={Step-DPO: Step-wise Preference Optimization for Long-chain Reasoning of LLMs},
    author={Xin Lai and Zhuotao Tian and Yukang Chen and Senqiao Yang and Xiangru Peng and Jiaya Jia},
    year={2024},
    eprint={2406.18629},
    archivePrefix={arXiv},
    primaryClass={cs.LG}
}

@misc{liu2024tisdpo,
    title={TIS-DPO: Token-level Importance Sampling for Direct Preference Optimization With Estimated Weights},
    author={Aiwei Liu and Haoping Bai and Zhiyun Lu and Yanchao Sun and Xiang Kong and Simon Wang and Jiulong Shan and Albin Madappally Jose and Xiaojiang Liu and Lijie Wen and Philip S. Yu and Meng Cao},
    year={2024},
    eprint={2410.04350},
    archivePrefix={arXiv},
    primaryClass={cs.CL}
}

@misc{wang20258020rulehighentropyminority,
      title={Beyond the 80/20 Rule: High-Entropy Minority Tokens Drive Effective Reinforcement Learning for LLM Reasoning}, 
      author={Shenzhi Wang and Le Yu and Chang Gao and Chujie Zheng and Shixuan Liu and Rui Lu and Kai Dang and Xionghui Chen and Jianxin Yang and Zhenru Zhang and Yuqiong Liu and An Yang and Andrew Zhao and Yang Yue and Shiji Song and Bowen Yu and Gao Huang and Junyang Lin},
      year={2025},
      eprint={2506.01939},
      archivePrefix={arXiv},
      primaryClass={cs.CL},
      url={https://arxiv.org/abs/2506.01939}, 
}

@misc{cui2025entropymechanismreinforcementlearning,
      title={The Entropy Mechanism of Reinforcement Learning for Reasoning Language Models}, 
      author={Ganqu Cui and Yuchen Zhang and Jiacheng Chen and Lifan Yuan and Zhi Wang and Yuxin Zuo and Haozhan Li and Yuchen Fan and Huayu Chen and Weize Chen and Zhiyuan Liu and Hao Peng and Lei Bai and Wanli Ouyang and Yu Cheng and Bowen Zhou and Ning Ding},
      year={2025},
      eprint={2505.22617},
      archivePrefix={arXiv},
      primaryClass={cs.LG},
      url={https://arxiv.org/abs/2505.22617}, 
}

@article{sprague2023musr,
  title={Musr: Testing the limits of chain-of-thought with multistep soft reasoning},
  author={Sprague, Zayne and Ye, Xi and Bostrom, Kaj and Chaudhuri, Swarat and Durrett, Greg},
  journal={arXiv preprint arXiv:2310.16049},
  year={2023}
}

@article{geva2021did,
  title={Did aristotle use a laptop? a question answering benchmark with implicit reasoning strategies},
  author={Geva, Mor and Khashabi, Daniel and Segal, Elad and Khot, Tushar and Roth, Dan and Berant, Jonathan},
  journal={Transactions of the Association for Computational Linguistics},
  volume={9},
  pages={346--361},
  year={2021},
  publisher={MIT Press One Rogers Street, Cambridge, MA 02142-1209, USA journals-info~…}
}

@article{jin2021disease,
  title={What disease does this patient have? a large-scale open domain question answering dataset from medical exams},
  author={Jin, Di and Pan, Eileen and Oufattole, Nassim and Weng, Wei-Hung and Fang, Hanyi and Szolovits, Peter},
  journal={Applied Sciences},
  volume={11},
  number={14},
  pages={6421},
  year={2021},
  publisher={MDPI}
}

@article{zhang2023evaluating,
  title={Evaluating the performance of large language models on gaokao benchmark},
  author={Zhang, Xiaotian and Li, Chunyang and Zong, Yi and Ying, Zhengyu and He, Liang and Qiu, Xipeng},
  journal={arXiv preprint arXiv:2305.12474},
  year={2023}
}
